\documentclass{ieeeaccess}
\usepackage{cite}
\usepackage{amsmath,amssymb,amsfonts}
\usepackage{algorithmic}
\usepackage{graphicx}
\usepackage{textcomp}
\usepackage{booktabs}
\usepackage{multirow}
\usepackage{makecell}
\usepackage{array}
\usepackage{caption}
\usepackage{wrapfig}
\usepackage{placeins}
\usepackage{soul}
\usepackage{xcolor}
\usepackage{float}
\usepackage{listings}

\def\BibTeX{{\rm B\kern-.05em{\sc i\kern-.025em b}\kern-.08em
    T\kern-.1667em\lower.7ex\hbox{E}\kern-.125emX}}
\begin{document}

\title{Candidate Generation and Definition-Guided Verification for Sentence-Level Depression Symptom Recognition}
\author{\uppercase{Weiming Li}\authorrefmark{1},
\uppercase{Catarina Barata}\authorrefmark{2},
\uppercase{Miguel Constante}\authorrefmark{3}, and
\uppercase{João Sanches}\authorrefmark{1}}
\address[1]{Institute for Systems and Robotics (ISR), LARSyS,
Departamento de Bioengenharia, Instituto Superior Técnico (IST), 1049-001 Lisboa. Portugal}
\address[2]{Institute for Systems and Robotics (ISR), LARSyS, Instituto Superior Técnico, University of Lisbon, 1049-001 Lisbon, Portugal}
\address[3]{Department of Psychiatry, Hospital Beatriz Ângelo, 2674-514 Loures, Portugal}

\tfootnote{
This work was supported by LARSyS funding
DOI:10.54499/LA/P/0083/2020 and 10.54499/UID/50009/2025.
This work has been submitted to the IEEE for possible publication.
Copyright may be transferred without notice, after which this version
may no longer be accessible.
}

\markboth
{Li \headeretal: Two-Stage Sentence-Level Depression Symptom Recognition}
{Li \headeretal: Two-Stage Sentence-Level Depression Symptom Recognition}

\corresp{Corresponding author: Weiming Li (E-mail: weimingli9999@gmail.com).}

\begin{abstract}
Sentence-level recognition of depression symptoms is challenging because similar expressions can differ in symptom relevance, and language-model inference is insufficiently grounded in diagnostic definitions. This study proposes a two-stage framework separating symptom-candidate generation from definition-grounded verification. A contrastively fine-tuned sentence encoder generates a symptom candidate per sentence, and a fine-tuned language model verifies whether the candidate is present or absent using the sentence, its context, and a candidate-specific diagnostic definition, checking its judgment against that definition before answering. Evaluated against encoder, inference-based, medical, and general LLM baselines and a matched single-stage supervised classifier, the proposed pipeline attains the best accuracy and F1 scores of all methods, with rationales matching expert-authored annotations. A preliminary clinical audit indicates moderate alignment with diagnostic definitions, with explanation quality strongly dependent on prediction correctness. The results support decomposing symptom recognition into candidate generation and definition-grounded verification, though performance remains limited for rare categories.
\end{abstract}


\titlepgskip=-15pt

\doi{}

\maketitle

\section{Introduction}
\label{sec:introduction}

The rapid growth of social media and online mental-health communities has produced large volumes of unstructured text describing users' emotions, behaviours, and psychological experiences \cite{b1}. Accordingly, text-based depression analysis has progressed from lexicon-based sentiment analysis and conventional machine-learning classifiers to transformer-based semantic models \cite{b2,b3}. Most existing studies, however, operate at the post or user level and produce aggregate outcomes such as depression status or severity. Although useful for large-scale screening, these formulations do not necessarily identify which sentence expresses a particular depressive symptom or how that expression corresponds to a symptom category defined by the Diagnostic and Statistical Manual of Mental Disorders, Fifth Edition (DSM-5) \cite{b11,b12}. Recent work has begun to align deep-learning depression detectors more directly with DSM-5 criteria, but such systems still classify at the document or post level rather than attributing individual sentences to specific symptom categories \cite{b41}.

Sentence-level symptom recognition provides a more fine-grained and traceable analytical target. ReDSM-5 supports this task by linking sentences from social-media posts to DSM-5-aligned depressive symptom annotations while retaining access to their source-post context \cite{b18}. Nevertheless, symptom expressions may be implicit, semantically overlapping, negated, temporally qualified, or dependent on surrounding text. The long-tailed distribution of symptom categories further complicates the recognition of less frequent expressions.

Contextual encoders such as Sentence-BERT and MentalBERT provide effective representations for sentence classification \cite{b21,b30}, while contrastive learning can organise the representation space using symptom-level supervision \cite{b31}. However, even strong encoders may produce ambiguous decisions near boundaries between overlapping symptom categories. Autoregressive large language models (LLMs) provide additional contextual modelling capacity but are substantially more expensive when applied to every sentence. Moreover, despite their instruction-following and structured-generation capabilities, LLMs remain subject to factual inconsistency, bias, prompt sensitivity, and imperfect agreement with clinicians \cite{b7,b8,b9,b10,b28}. Sentence-level DSM-5 recognition therefore requires both selective use of LLM reasoning and an explicit reference for determining whether a predicted symptom is supported by the textual evidence. Prior sentence-level approaches, however, either rely on a single encoder-only classifier without a definitional consistency check, apply generic zero-shot entailment that is not conditioned on DSM-5 criteria, or invoke LLM reasoning on every sentence without separating candidate generation from verification \cite{b18,b34,b24,b25}. To our knowledge, prior work has not examined this specific combination of encoder-based candidate generation and candidate-specific, definition-grounded verification, in which LLM verification is restricted to the selected symptom candidate.

To address these challenges, we propose a two-stage sentence-level DSM-5 symptom recognition framework that separates symptom-candidate generation from candidate-specific presence verification. In Stage~1, MentalSBERT-S, a sentence encoder contrastively fine-tuned with symptom-level supervision, produces a nine-class symptom-matching score for each sentence and selects the highest-scoring category as the Stage-1 candidate. This encoder-only design avoids autoregressive generation for every sentence while matching the accuracy of the directly fine-tuned generative alternative, as shown in Section~IV.

In Stage~2, a separately fine-tuned DeepSeek model determines whether the Stage-1 candidate is \textsc{Present} or \textsc{Absent}. It receives the target sentence, source-post context, candidate symptom, and a concise candidate-specific DSM-5-informed definition. Within one structured inference call, the model forms a binary judgment and checks its consistency with the supplied definition before answering. The framework consequently produces a sentence-level symptom hypothesis, a consistency-checked binary status, and a model-generated explanation grounded in the target sentence and its supporting context. This verification operates at the annotation level and is not intended as a complete clinical diagnostic assessment.

The main contributions of this work are as follows:

\begin{enumerate}
    \item We formulate sentence-level DSM-5 symptom recognition as a two-stage task that separates nine-class symptom-candidate generation from definition-grounded binary verification of the selected candidate.

    \item We develop MentalSBERT-S, a contrastively fine-tuned sentence encoder that generates the Stage-1 symptom candidate directly from a single forward pass, matching the accuracy of the directly fine-tuned generative alternative without autoregressive inference.

    \item We introduce a fine-tuned definition-grounded verification procedure that combines an initial \textsc{Present}/\textsc{Absent} judgment with candidate-specific, DSM-5-informed consistency checking within one structured inference call.

    \item We evaluate the framework against supervised encoders, NLI-style zero-shot models, direct LLM baselines including MedGemma-4B-IT, and a matched single-stage supervised classifier trained on the identical data and backbone. The evaluation includes three training seeds, hierarchical bootstrap confidence intervals, per-symptom analysis, component ablations, error analysis, inference-efficiency measurements, and a preliminary clinical rationale audit.
\end{enumerate}

The remainder of this paper is organised as follows. Section~II reviews related work. Section~III presents the proposed framework. Section~IV describes the experimental setup and results. Section~V discusses the findings, clinical implications, and limitations. Section~VI concludes the paper and outlines future research directions.

\section{Related Work}
\label{sec:related_work}

Early text-based depression detection primarily addressed user-level or post-level screening through binary classification or severity estimation \cite{b11,b12,b5}. Resources such as SMHD, expert-annotated suicide-risk posts, interview corpora, and contextual depression challenges have broadened the empirical foundations of mental-health NLP \cite{b14,b15,b16,b17}. However, predictions aggregated across a document or user history do not necessarily localise the sentence expressing a clinically relevant symptom \cite{b13}. ReDSM-5 advances a more fine-grained formulation by associating individual sentences with DSM-5-aligned depressive symptom categories \cite{b18}. In this setting, the linked post remains useful for resolving negation, temporal qualification, and unclear references, but the prediction target is the symptom evidence expressed by an individual sentence.

Pretrained contextual encoders remain strong foundations for mental-health text analysis. Domain adaptation to mental-health corpora can improve representations of language associated with mental-health conditions \cite{b19,b20}. MentalBERT and MentalRoBERTa, for example, were pretrained on mental-health-related social-media text and evaluated across multiple mental-health detection benchmarks \cite{b21}. MentalBERT representations have also been combined with downstream neural classifiers for depression and major-depressive-disorder detection \cite{b22}. Transfer learning from general-purpose encoders to subreddit-specific mental-illness classification has likewise been shown to improve label discrimination over generic pretraining \cite{b42}. Although these models can learn symptom-discriminative contextual representations, fixed-label decisions remain difficult when symptom categories share linguistic cues, evidence depends on surrounding context, or rare categories provide limited supervision. Sentence-pair and contrastive objectives can further shape the representation space by drawing same-label sentences together and separating cross-category sentences \cite{b30,b31}; nevertheless, their contribution must be established empirically through controlled ablation.

Natural-language inference (NLI) offers a complementary formulation in which label descriptions are expressed as hypotheses and evaluated for entailment. Yin et al.\ \cite{b34} established entailment as a general approach to zero-shot text classification, while BART provides a widely used sequence-to-sequence foundation for NLI-style transfer \cite{b35}. This formulation is relevant to symptom recognition because symptom definitions can be converted into hypotheses without task-specific fine-tuning. However, generic entailment training does not necessarily capture the annotation boundaries of a specialised mental-health dataset. We therefore treat NLI-style zero-shot classification as an external baseline rather than a replacement for supervised symptom modelling.

LLMs have expanded the range of clinical and mental-health NLP tasks that can be addressed through instruction following, contextual reasoning, and structured generation \cite{b24,b25}. A recent systematic review of LLM use in mental health similarly reports that most deployments remain confined to screening-level classification and stresses the need for interpretable, clinically verifiable outputs rather than opaque predictions \cite{b43}. Medical instruction tuning and domain alignment can improve performance on medical tasks, but strong benchmark results do not eliminate concerns regarding possible harm, bias, prompt sensitivity, or disagreement with clinicians \cite{b26,b28}. These concerns are especially relevant when heterogeneous expressions of symptoms such as anhedonia must be mapped to narrowly defined clinical categories \cite{b23}. Direct zero-shot and few-shot LLM methods are therefore important baselines, while constrained outputs and explicit separation between model-generated explanations and clinically reviewed interpretations remain necessary.

Medically tuned foundation models provide a stronger comparison than general-purpose LLM prompting alone. MedGemma is a family of medical text and vision-language models derived from Gemma~3 and evaluated across medical reasoning, information-retrieval, and image-understanding tasks \cite{b37}. MedGemma-4B-IT is an instruction-tuned model of practical size that can be applied to textual medical tasks without dataset-specific training. Its inclusion as a zero-shot baseline tests whether general medical instruction tuning is sufficient to capture the sentence-level DSM-5 annotation boundaries considered in this study.

Clinical knowledge can be incorporated into LLM inference through diagnostic definitions, structured instructions, external documents, or retrieval-augmented generation \cite{b27,b28,b29}. Retrieval-oriented systems introduce external evidence, whereas definition-conditioned inference supplies a fixed criterion directly in the model input. For sentence-level DSM-5 recognition, concise candidate-specific definitions provide a transparent reference for assessing whether the target sentence and its local context support a proposed symptom. Such definition-grounded, self-checking verification remains an annotation-level consistency check rather than a complete diagnostic assessment.

Collectively, prior work provides effective components for contextual encoding, contrastive representation learning, zero-shot entailment, medical instruction following, and definition-conditioned reasoning. However, these components address different parts of the sentence-level recognition problem. The present study integrates them into a two-stage pipeline with separately evaluated responsibilities: encoder-based symptom-candidate generation and definition-grounded binary verification of the resulting candidate.

\section{Method}
\label{sec:method}

We propose a two-stage sentence-level DSM-5 symptom recognition framework that separates symptom-category localisation from symptom-presence verification. The design follows a ``semantic filtering first, structured verification second'' principle. Stage~1 answers \emph{which symptom category best describes the target sentence}; Stage~2 answers \emph{whether that candidate symptom is actually supported}. These responsibilities are separated because category confusion and symptom validity are different error sources. A sentence can be semantically close to a symptom category without providing sufficient evidence that the symptom is actually present, particularly when the expression is negated or context-dependent. The framework therefore keeps candidate selection and evidence verification as distinct decisions. It remains sentence-level: the target sentence is the unit being labelled, while its linked post is used only as supporting context when needed.

\begin{table*}[!t]
\centering
\caption{Example post from the ReDSM-5 dataset~\cite{b18} illustrating the sentence-level annotation format used throughout this study. The full source post (left) provides context; three of its sentences (second column) are each independently labelled with a DSM-5 symptom category and a \textsc{Present}/\textsc{Absent} status (third column), together with a human-written rationale explaining the label (right column). This structure motivates the sentence-level, definition-grounded formulation adopted in Section~\ref{sec:method}.}
\label{tab:dataset_example}
\renewcommand{\arraystretch}{1.15}
\setlength{\tabcolsep}{4pt}
\begin{tabular}{>{\centering\arraybackslash}p{0.20\textwidth} >{\centering\arraybackslash}p{0.20\textwidth} >{\centering\arraybackslash}p{0.16\textwidth} >{\centering\arraybackslash}p{0.34\textwidth}}
\toprule
\multicolumn{1}{c}{\textbf{Post text}} &
\multicolumn{1}{c}{\textbf{Sentence text}} &
\multicolumn{1}{c}{\textbf{DSM-5 symptom}} &
\multicolumn{1}{c}{\textbf{Annotation explanation}} \\
\midrule
\multirow{3}{0.20\textwidth}{
\centering
I felt really good when I first started it.\\
(I was filled with energy).\\
I got so much done for 2--3 months.\\
It all went downhill after that\\
(I can't concentrate at all now).\\
(I have more sleep troubles and now rely on sleeping pills.\\
Also in constant low moods)
}
& \centering Also in constant low moods
& \centering DEPRESSED\_MOOD\\\textsc{Present}
& \centering The sentence directly expresses a continuing low mood, which is consistent with depressed-mood symptom evidence at the sentence level.
\tabularnewline
\cmidrule(l){2-4}
& \centering I can't concentrate at all now
& \centering COGNITIVE\_ISSUES\\\textsc{Present}
& \centering Difficulty concentrating or thinking clearly is a hallmark of depression, aligning with criterion eight of the DSM-5 major depressive disorder symptoms.
\tabularnewline
\cmidrule(l){2-4}
& \centering I have more sleep troubles and now rely on sleeping pills
& \centering SLEEP\_ISSUES\\\textsc{Present}
& \centering Persistent sleep difficulty and reliance on medication may indicate sleep disturbance, corresponding to insomnia or hypersomnia within the DSM-5 symptom set.
\tabularnewline
\bottomrule
\end{tabular}
\end{table*}

This study uses the ReDSM-5 dataset~\cite{b18}, which links each target sentence to its source post and provides a primary symptom category. The sentence is the minimal analytical unit, while the linked post provides supporting context for LLM-based verification when sentence meaning depends on surrounding text. Table~\ref{tab:dataset_example} illustrates this organisation. Model-parameter training and training-pair construction use only the training partition.

\begin{figure}[!t]
\centering
\includegraphics[width=\columnwidth,height=0.45\textheight,keepaspectratio]{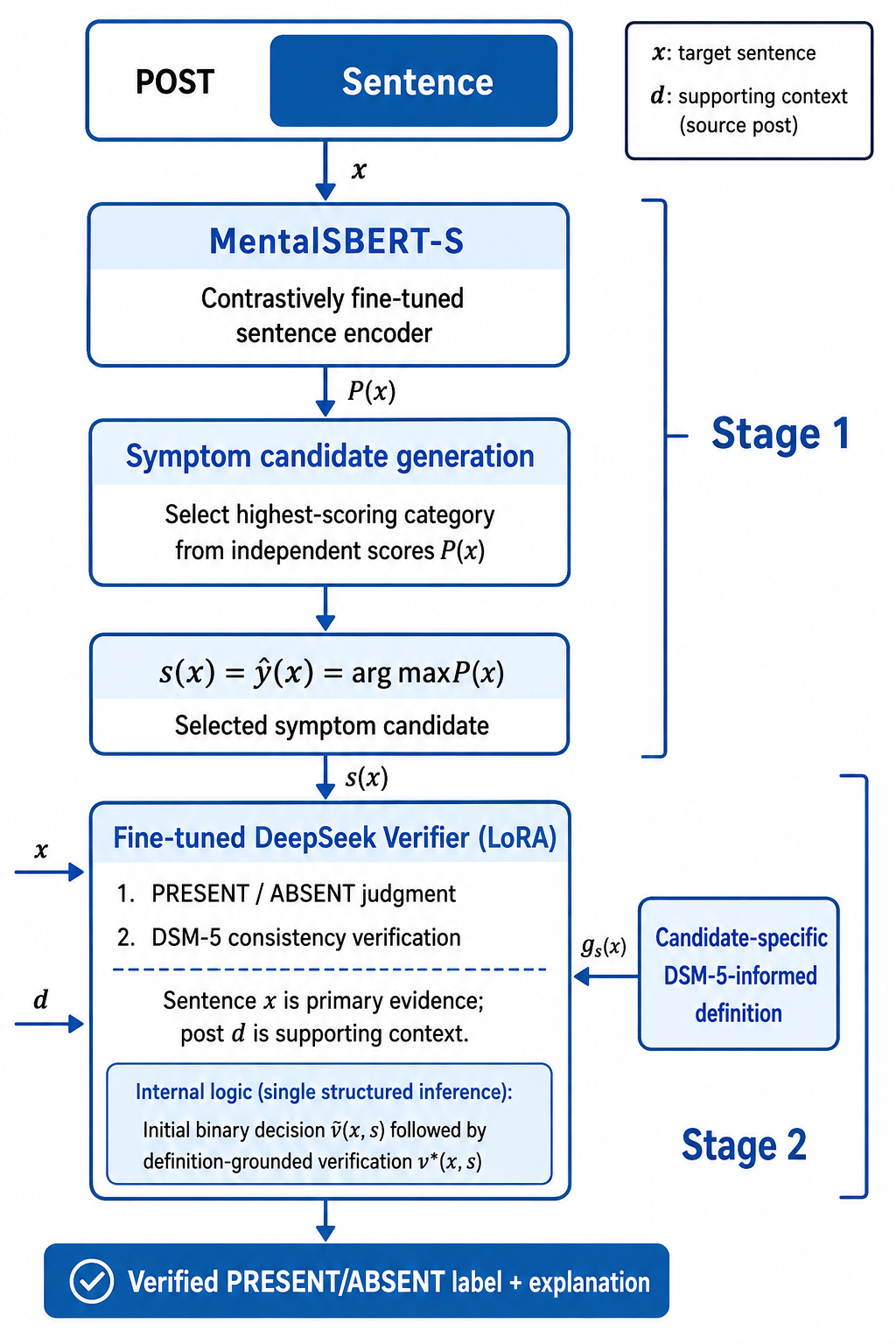}
\caption{Two-stage sentence-level symptom recognition framework. Stage~1 generates a symptom candidate using a contrastively fine-tuned sentence encoder, and Stage~2 performs binary symptom-presence judgment with definition-grounded, internal self-checking against the candidate-specific DSM-5-informed definition, all within a single model call.}
\label{fig:framework}
\end{figure}

As illustrated in Fig.~\ref{fig:framework}, the framework separates symptom-candidate generation from presence verification. Given a target sentence $x$ and source post $d$, Stage~1 uses MentalSBERT-S to produce independent symptom scores $P(x)$ and selects the highest-scoring category as the candidate $s(x)=\hat{y}(x)$. Stage~2 provides $x$, $d$, $s(x)$, and the corresponding candidate-specific DSM-5-informed definition $g_{s(x)}$ to a separately fine-tuned DeepSeek verifier, which determines whether the candidate is \textsc{Present} or \textsc{Absent} and checks consistency with the definition in a single structured inference call. The target sentence remains the primary evidence, while the post supplies supporting context for ambiguity, negation, temporal information, or context-dependent expressions. The final output is the verified status together with a model-generated explanation.

\subsection{Symptom-Aware Sentence Representation and Initial Classification}
\label{subsec:stage1_encoder}

Let $x$ denote a target sentence and $d$ its source post. Stage~1 formulates symptom-candidate generation as a single-label classification problem over
\begin{equation}
\mathcal{Y}=\{1,2,\ldots,K\}, \qquad K=9,
\label{eq:label_space}
\end{equation}
where each label corresponds to one of the nine core depressive symptom categories represented in ReDSM-5.

MentalSBERT-S is initialised from a SentenceTransformer encoder~\cite{b30} and fine-tuned on the ReDSM-5 training split. Given contextual token representations $z_i\in\mathbb{R}^{m}$ for a sentence containing $n$ subword tokens, mean pooling produces
\begin{equation}
h(x)=\frac{1}{n}\sum_{i=1}^{n}z_i,
\label{eq:mean_pooling}
\end{equation}
where $h(x)\in\mathbb{R}^{m}$ is the sentence embedding. The encoder is trained using symptom supervision together with label-derived contrastive relations. Positive relations are formed only between training sentences sharing the same symptom label, whereas sentences with different labels provide cross-category negative relations. No held-out sentence is used to construct these relations.

On the shared representation $h(x)$, the model uses $K$ one-vs-rest classification heads. For label $k$, the independent sigmoid score is
\begin{equation}
p_k(x)=\sigma\!\left(w_k^{\top}h(x)+b_k\right), \qquad k\in\mathcal{Y}.
\label{eq:independent_score}
\end{equation}
The vector
\begin{equation}
P(x)=\left[p_1(x),p_2(x),\ldots,p_K(x)\right]
\label{eq:confidence_vector}
\end{equation}
contains independently estimated class-matching scores rather than a jointly normalised categorical posterior. The initial symptom prediction is
\begin{equation}
\hat{y}(x)=\arg\max_{k\in\mathcal{Y}}p_k(x).
\label{eq:prediction}
\end{equation}
Independent heads allow overlapping symptoms to receive non-negligible scores without forcing probability mass to sum to one. The Stage-1 symptom candidate is $s(x)=\hat{y}(x)$, which is passed directly to Stage~2 for verification.

\subsection{Definition-Grounded Binary Symptom Verification}
\label{subsec:headB}

Stage~2 verifies the candidate symptom $s(x)$ produced by Stage~1. It is formulated as binary classification rather than another nine-class prediction:
\begin{equation}
v(x,s(x))\in\{0,1\},
\label{eq:binary_status}
\end{equation}
where $1$ denotes \textsc{Present} and $0$ denotes \textsc{Absent}. The target remains the sentence $x$; its source post $d$ is included only to resolve contextual phenomena such as negation, referent ambiguity, or statements whose interpretation depends on surrounding text.

The verifier is a DeepSeek decoder separately adapted by low-rank fine-tuning (LoRA)~\cite{b32} for this present/absent task. Its training objective is
\begin{equation}
\mathcal{L}_{B}
=-\sum_{t=1}^{T_B}\log p_{\theta_B}
\left(u_t\mid u_{<t},d,x,s(x)\right),
\label{eq:headB_loss}
\end{equation}
where $u_t$ is a target output token, $T_B$ is the length of the target output sequence, and $\theta_B$ denotes the verifier's LoRA-adapted parameters.

Given the target sentence, post context, and Stage-1 candidate, the verifier records an initial binary judgment
\begin{equation}
\tilde v(x,d,s(x))=f_B\!\left(d,x,s(x)\right).
\label{eq:initial_binary}
\end{equation}
The inference prompt instructs the model to prioritise explicit evidence in $x$, use $d$ only as supporting context, distinguish current symptom evidence from negated or historical-only statements, and return a structured JSON object. The model is not asked to diagnose major depressive disorder, and the binary label does not require evidence for all diagnostic criteria, duration, severity, or functional impairment.

\subsection{Definition-Grounded Self-Checking and Explanation Generation}
\label{subsec:dsm_verify}

For each candidate symptom $s(x)$, a concise DSM-5-informed symptom definition $g_{s(x)}$ is supplied as a sentence-level consistency reference. This definition guides verification of the candidate symptom but does not constitute a complete diagnostic assessment of duration, severity, functional impairment, or the full DSM-5 criteria for major depressive disorder. The model checks whether its preliminary binary decision is semantically consistent with the target sentence, supporting context, and candidate-specific definition, producing
\begin{equation}
v^{*}(x,d,s(x))=V_{\mathrm{DSM5}}
\left(d,x,s(x),\tilde v(x,d,s(x)),g_{s(x)}\right).
\label{eq:dsm_verify}
\end{equation}
The initial judgment and consistency check are elicited as two structured fields within one verifier model call. Equation~\eqref{eq:dsm_verify} describes the internal decision logic and does not denote a separate model or a second inference call. A second call is made only if syntactically invalid JSON requires repair.

\section{Experimental Results}
\label{sec:experiments}

This section describes the dataset and experimental protocol and then evaluates Stage-1 symptom localisation, Stage-2 binary verification, and the complete two-stage framework. The analysis covers comparisons with supervised encoder, NLI-style, medical-LLM, and direct LLM baselines, together with statistical robustness, computational efficiency, error patterns, and automatic and psychiatrist-based assessment of model-generated explanations.

\subsection{Dataset}
\label{subsec:dataset}

This study uses ReDSM-5, a sentence-level dataset derived from DepreSym~\cite{b39} by linking annotated sentences to their original Reddit posts and assigning DSM-5-informed symptom labels~\cite{b18}. Each record contains a target sentence, source-post context, one symptom label, a candidate-specific \textsc{Present}/\textsc{Absent} status, and an annotation rationale. The label space covers nine core depressive symptom categories, supporting multiclass symptom localisation in Stage~1 and candidate-specific binary verification in Stage~2; Table~\ref{tab:dsm5_definitions} summarises these categories and definitions~\cite{b18,b44}. The raw corpus contains 1,896 uniquely annotated sentences. Before partitioning, we apply two cleaning steps. First, we remove 82 sentences labelled only as \texttt{SPECIAL\_CASE}, an operational residual category for psychologically relevant content outside the nine core DSM-5 symptoms. Second, we remove 47 genuinely multi-label sentences annotated as simultaneously \textsc{Present} for two or more categories, since assigning a single primary label would impose an arbitrary priority rather than reflect the annotation. The resulting dataset contains 1,767 single-label sentences across the nine core categories.

Because a Reddit post may contribute multiple annotated sentences, sentence-level splitting can leak the same post across partitions. We therefore use CP-SAT~\cite{b40} to assign whole posts to train, development, and test sets while minimizing deviations from a 65/16/19\% per-class target. This yields 1,163/280/324 instances with zero post-level overlap. The prediction unit is the annotated target sentence, with its source post providing context for negation, temporal qualification, unclear referents, or information distributed across nearby sentences. Labels represent sentence--symptom correspondence rather than patient-level diagnosis, so diagnostic duration, frequency, severity, impairment, and differential-diagnosis criteria are not required. Table~\ref{tab:dsm5_definitions} is therefore used as a semantic reference for symptom-related evidence, not as a complete diagnostic rule set. All parameter training, contrastive-pair construction, fine-tuning, and checkpoint optimisation use only the training partition; the development partition is used for model selection and early stopping. The test partition is touched only once, for final evaluation. Evaluation labels and annotation rationales are not supplied to any model during inference.

\par\medskip

\noindent

\begin{minipage}{\columnwidth}

\captionsetup{
    justification=centering,
    singlelinecheck=false
}

\captionof{table}{
DSM-5-Informed Symptom Categories and Definitions~\cite{b18,b44}
}
\label{tab:dsm5_definitions}

\vspace{0.3em}

\centering

\renewcommand{\arraystretch}{1.08}
\setlength{\tabcolsep}{3pt}
\footnotesize

\begin{tabular}{
    >{\centering\arraybackslash}p{0.30\columnwidth}
    >{\centering\arraybackslash}p{0.63\columnwidth}
}
\toprule

\multicolumn{1}{c}{\textbf{Symptom Category}} &
\multicolumn{1}{c}{\textbf{Definition}}
\tabularnewline

\midrule

\textbf{Depressed Mood}
&
Depressed mood most of the day, nearly every day, as indicated
by either subjective report (e.g., feels sad, empty, or hopeless)
or observation made by others (e.g., appears tearful).
\tabularnewline

\midrule

\textbf{Anhedonia}
&
Markedly diminished interest or pleasure in all, or almost all,
activities most of the day, nearly every day (as indicated by
either subjective account or observation).
\tabularnewline

\midrule

\textbf{Appetite or Weight Change}
&
Significant weight loss when not dieting or weight gain
(e.g., a change of more than 5\% of body weight in a month),
or decrease or increase in appetite nearly every day.
\tabularnewline

\midrule

\textbf{Sleep Issues}
&
Insomnia or hypersomnia nearly every day.
\tabularnewline

\midrule

\textbf{Psychomotor Alteration}
&
Psychomotor agitation or retardation nearly every day
(observable by others, not merely subjective feelings of
restlessness or being slowed down).
\tabularnewline

\midrule

\textbf{Fatigue / Loss of Energy}
&
Fatigue or loss of energy nearly every day.
\tabularnewline

\midrule

\textbf{Worthlessness or Excessive Guilt}
&
Feelings of worthlessness or excessive or inappropriate guilt
(which may be delusional) nearly every day (not merely
self-reproach or guilt about being sick).
\tabularnewline

\midrule

\textbf{Cognitive Issues}
&
Diminished ability to think or concentrate, or indecisiveness,
nearly every day (either by subjective account or as observed
by others).
\tabularnewline

\midrule

\textbf{Suicidal Thoughts / Death Ideation}
&
Recurrent thoughts of death (not just fear of dying), recurrent
suicidal ideation without a specific plan, or a suicide attempt
or a specific plan for committing suicide.
\tabularnewline

\bottomrule
\end{tabular}

\end{minipage}

\par\medskip

\begin{table*}[!t]
\centering
\captionsetup{justification=centering}
\caption{Stage-1 per-symptom precision, recall, and F1-score comparison for ANH=ANHEDONIA; APP=APPETITE\_CHANGE; COG=COGNITIVE\_ISSUES; DEP=DEPRESSED\_MOOD; FAT=FATIGUE. \textit{Note:} For methods evaluated over three seeds, each entry reports the seed-averaged precision/recall/F1; deterministic zero-shot, few-shot, and retrieval-augmented methods are evaluated once.}
\label{tab:stage1_prf_part1}
\footnotesize
\renewcommand{\arraystretch}{1.15}
\setlength{\tabcolsep}{3.8pt}
\resizebox{\textwidth}{!}{
\begin{tabular}{cccccc}
\toprule
\multicolumn{1}{c}{\textbf{Model}} &
\multicolumn{1}{c}{\textbf{ANH}} &
\multicolumn{1}{c}{\textbf{APP}} &
\multicolumn{1}{c}{\textbf{COG}} &
\multicolumn{1}{c}{\textbf{DEP}} &
\multicolumn{1}{c}{\textbf{FAT}} \\
\midrule
BART-MNLI zero-shot & \makecell{0.266/0.943/0.415} & \makecell{1.000/0.083/0.154} & \makecell{0.400/0.095/0.154} & \makecell{0.675/0.333/0.446} & \makecell{0.450/0.818/0.581} \\
MedGemma-4B-IT zero-shot & \makecell{0.595/0.714/0.649} & \makecell{0.889/0.667/0.762} & \makecell{0.812/0.619/0.703} & \makecell{0.577/0.741/0.649} & \makecell{0.718/0.848/0.778} \\
MentalBERT zero-shot & \makecell{0.771/0.771/0.771} & \makecell{1.000/1.000/1.000} & \makecell{0.643/0.857/0.735} & \makecell{0.853/0.358/0.504} & \makecell{0.882/0.909/0.895} \\
DeBERTa & \makecell{0.678/0.886/0.758} & \makecell{0.000/0.000/0.000} & \makecell{0.286/0.095/0.143} & \makecell{0.887/0.831/0.857} & \makecell{0.465/0.707/0.557} \\
MentalRoBERTa & \makecell{0.893/0.924/0.907} & \makecell{1.000/0.944/0.971} & \makecell{0.920/0.905/0.912} & \makecell{0.931/0.938/0.935} & \makecell{0.806/1.000/0.892} \\
DeepSeek 0-shot & \makecell{0.698/0.857/0.769} & \makecell{1.000/0.833/0.909} & \makecell{0.615/0.762/0.681} & \makecell{0.768/0.778/0.773} & \makecell{0.912/0.939/0.925} \\
DeepSeek few-shot & \makecell{0.566/0.857/0.682} & \makecell{1.000/0.917/0.957} & \makecell{0.652/0.714/0.682} & \makecell{0.879/0.630/0.734} & \makecell{0.879/0.879/0.879} \\
DeepSeek RAG & \makecell{0.630/0.829/0.716} & \makecell{1.000/0.667/0.800} & \makecell{0.889/0.762/0.821} & \makecell{0.757/0.691/0.723} & \makecell{0.882/0.909/0.895} \\
DeepSeek (fine-tuned) & \makecell{0.923/0.905/0.914} & \makecell{1.000/0.889/0.938} & \makecell{0.903/0.873/0.887} & \makecell{0.895/0.942/0.918} & \makecell{0.838/0.990/0.908} \\
MentalSBERT-S (BCE only) & \makecell{0.930/0.886/0.907} & \makecell{1.000/0.972/0.986} & \makecell{0.944/0.810/0.870} & \makecell{0.903/0.951/0.926} & \makecell{0.860/0.990/0.920} \\
\textbf{MentalSBERT-S} & \makecell{\textbf{0.969/0.895/0.930}} & \makecell{\textbf{1.000/0.917/0.957}} & \makecell{\textbf{0.945/0.825/0.881}} & \makecell{\textbf{0.912/0.975/0.942}} & \makecell{\textbf{0.831/0.990/0.903}} \\
\bottomrule
\end{tabular}}
\end{table*}

\begin{table*}[!t]
\centering
\captionsetup{justification=centering}
\caption{Stage-1 per-symptom precision, recall, and F1-score comparison for PSY=PSYCHOMOTOR; SLE=SLEEP\_ISSUES; SUI=SUICIDAL\_THOUGHTS; WOR=WORTHLESSNESS. \textit{Note:} For methods evaluated over three seeds, each entry reports the seed-averaged precision/recall/F1; deterministic zero-shot, few-shot, and retrieval-augmented methods are evaluated once.}
\label{tab:stage1_prf_part2}
\footnotesize
\renewcommand{\arraystretch}{1.15}
\setlength{\tabcolsep}{3.8pt}
\resizebox{0.75\textwidth}{!}{
\begin{tabular}{ccccc}
\toprule
\multicolumn{1}{c}{\textbf{Model}} &
\multicolumn{1}{c}{\textbf{PSY}} &
\multicolumn{1}{c}{\textbf{SLE}} &
\multicolumn{1}{c}{\textbf{SUI}} &
\multicolumn{1}{c}{\textbf{WOR}} \\
\midrule
BART-MNLI zero-shot & \makecell{0.117/0.778/0.203} & \makecell{1.000/0.194/0.324} & \makecell{1.000/0.771/0.871} & \makecell{1.000/0.015/0.029} \\
MedGemma-4B-IT zero-shot & \makecell{0.125/0.111/0.118} & \makecell{0.889/0.774/0.828} & \makecell{1.000/0.800/0.889} & \makecell{0.938/0.672/0.783} \\
MentalBERT zero-shot & \makecell{0.600/0.333/0.429} & \makecell{0.906/0.935/0.921} & \makecell{0.557/0.971/0.708} & \makecell{0.554/0.687/0.613} \\
DeBERTa & \makecell{0.000/0.000/0.000} & \makecell{0.733/0.925/0.802} & \makecell{0.927/0.952/0.939} & \makecell{0.839/0.905/0.870} \\
MentalRoBERTa & \makecell{0.833/0.333/0.474} & \makecell{0.966/0.925/0.945} & \makecell{0.941/0.914/0.928} & \makecell{0.950/0.930/0.940} \\
DeepSeek 0-shot & \makecell{0.167/0.111/0.133} & \makecell{0.935/0.935/0.935} & \makecell{0.941/0.914/0.927} & \makecell{0.862/0.746/0.800} \\
DeepSeek few-shot & \makecell{0.176/0.333/0.231} & \makecell{0.903/0.903/0.903} & \makecell{0.968/0.857/0.909} & \makecell{0.862/0.836/0.849} \\
DeepSeek RAG & \makecell{0.400/0.444/0.421} & \makecell{1.000/0.935/0.967} & \makecell{0.689/0.886/0.775} & \makecell{0.863/0.657/0.746} \\
DeepSeek (fine-tuned) & \makecell{0.838/0.519/0.637} & \makecell{0.935/0.925/0.930} & \makecell{0.980/0.933/0.956} & \makecell{0.954/0.920/0.937} \\
MentalSBERT-S (BCE only) & \makecell{1.000/0.333/0.477} & \makecell{0.957/0.957/0.957} & \makecell{0.961/0.924/0.942} & \makecell{0.924/0.965/0.944} \\
\textbf{MentalSBERT-S} & \makecell{\textbf{0.867/0.444/0.586}} & \makecell{\textbf{0.957/0.946/0.951}} & \makecell{\textbf{1.000/0.943/0.971}} & \makecell{\textbf{0.956/0.970/0.963}} \\
\bottomrule
\end{tabular}}
\end{table*}

\begin{table*}[!t]
\centering
\captionsetup{justification=centering}
\caption{
Overall Stage-1 performance on the 324-instance test partition.
\newline
\footnotesize
\textit{Note:}
For methods evaluated over three seeds, each entry reports the seed mean$\pm$sample SD.
Methods evaluated once (deterministic zero-shot, few-shot, and retrieval-augmented
prompting) omit SD.
}
\label{tab:overall_metrics}

\scriptsize
\renewcommand{\arraystretch}{1.05}
\setlength{\tabcolsep}{3.5pt}

\begin{tabular}{lccc}
\toprule

\textbf{Model} &
\textbf{Accuracy} &
\textbf{Macro F1} &
\textbf{Weighted F1}
\\

\midrule

BART-MNLI zero-shot & 0.404 & 0.353 & 0.368 \\
MedGemma-4B-IT zero-shot & 0.716 & 0.684 & 0.726 \\
MentalBERT zero-shot & 0.704 & 0.731 & 0.689 \\
DeBERTa & 0.760$\pm$0.035 & 0.547$\pm$0.025 & 0.720$\pm$0.032 \\
MentalRoBERTa & 0.919$\pm$0.011 & 0.878$\pm$0.010 & 0.916$\pm$0.011 \\
DeepSeek 0-shot & 0.809 & 0.762 & 0.807 \\
DeepSeek few-shot & 0.781 & 0.758 & 0.793 \\
DeepSeek RAG & 0.762 & 0.763 & 0.774 \\
DeepSeek (fine-tuned) & 0.918$\pm$0.010 & 0.891$\pm$0.020 & 0.916$\pm$0.010 \\
MentalSBERT-S (BCE only) & 0.923$\pm$0.003 & 0.881$\pm$0.024 & 0.918$\pm$0.005 \\
\textbf{MentalSBERT-S} & \textbf{0.934$\pm$0.002} & \textbf{0.898$\pm$0.002} & \textbf{0.932$\pm$0.002} \\

\bottomrule
\end{tabular}

\end{table*}

\begin{table*}[!t]
\centering
\captionsetup{justification=centering}
\caption{
Efficiency comparison across the retained Stage-1 baselines and ablations on the 324-instance test partition.
\newline
\footnotesize
\textit{Note:}
LLM Gen is the number of autoregressive generations. Multi-seed configurations report mean$\pm$SD across seeds.
}
\label{tab:efficiency_comparison}

\scriptsize
\renewcommand{\arraystretch}{1.08}
\setlength{\tabcolsep}{3pt}

\resizebox{\textwidth}{!}{
\begin{tabular}{cccccccc}
\toprule
\multicolumn{1}{c}{\textbf{Model}} &
\multicolumn{1}{c}{\textbf{LLM Gen}} &
\multicolumn{1}{c}{\textbf{Total Time (s)}} &
\multicolumn{1}{c}{\textbf{Avg Latency (s)}} &
\multicolumn{1}{c}{\textbf{Throughput}} &
\multicolumn{1}{c}{\textbf{Avg Input Tok.}} &
\multicolumn{1}{c}{\textbf{Avg Output Tok.}} &
\multicolumn{1}{c}{\textbf{Total Tokens}} \\
\midrule
BART-MNLI zero-shot & 0 & 3.35 & 0.010 & 96.65 & 311.1 & 0.0 & 100791 \\
MedGemma-4B-IT zero-shot & 324 & 156.4 & 0.483 & 2.07 & 426.7 & 13.7 & 142683 \\
MentalBERT zero-shot & 0 & 0.20 & 0.001 & 1639.26 & 18.3 & 0.0 & 5923 \\
DeBERTa & 0$\pm$0 & 1.11$\pm$0.02 & 0.004$\pm$0.000 & 251.92$\pm$3.38 & 316.8$\pm$0.0 & 0.0$\pm$0.0 & 88697$\pm$0 \\
MentalRoBERTa & 0$\pm$0 & 0.41$\pm$0.00 & 0.001$\pm$0.000 & 796.54$\pm$7.53 & 284.6$\pm$0.0 & 0.0$\pm$0.0 & 92195$\pm$0 \\
DeepSeek 0-shot & 324 & 3240.4 & 10.001 & 0.10 & 400.8 & 295.3 & 225527 \\
DeepSeek few-shot & 324 & 3965.6 & 12.240 & 0.08 & 973.1 & 342.0 & 426094 \\
DeepSeek RAG & 324 & 1823.5 & 5.628 & 0.18 & 599.4 & 145.4 & 241332 \\
DeepSeek (fine-tuned) & 324$\pm$0 & 292.4$\pm$2.0 & 0.902$\pm$0.006 & 1.11$\pm$0.01 & 242.9$\pm$3.5 & 11.9$\pm$0.2 & 82531$\pm$1100 \\
\textbf{MentalSBERT-S} & \textbf{0$\pm$0} & \textbf{0.39$\pm$0.10} & \textbf{0.001$\pm$0.000} & \textbf{869.75$\pm$208.77} & \textbf{18.3$\pm$0.0} & \textbf{0.0$\pm$0.0} & \textbf{5923$\pm$0} \\
\bottomrule
\end{tabular}}
\end{table*}

\subsection{Stage-1 Results Analysis}

Stage-1 assigns one of nine DSM-5 symptom categories to each target sentence. MentalSBERT-S is a SentenceTransformer encoder fine-tuned with a joint multi-head binary classification and triplet-contrastive objective. Methods with trainable multi-seed configurations were evaluated with random seeds 42, 52, and 62 on the fixed 324-sentence test partition, with results reported as the seed mean and sample standard deviation. Fixed single-run zero-shot, few-shot, and retrieval-augmented baseline configurations were evaluated once. BART-MNLI uses independent entailment scoring with a fixed hypothesis template, while generative LLM baselines use greedy decoding. MedGemma-4B-IT provides a recent medical-domain LLM baseline, evaluated under zero-shot prompting with the same task instruction and label set as the other generative baselines. MentalBERT zero-shot is a deterministic cosine-similarity classifier that embeds the target sentence and each symptom's textual definition with the same SentenceTransformer backbone used by MentalSBERT-S, without any task-specific training.

Tables~\ref{tab:stage1_prf_part1} and~\ref{tab:stage1_prf_part2} summarize class-specific performance, whereas Table~\ref{tab:overall_metrics} reports aggregate performance. Per-symptom cells report one-versus-rest precision, recall, and F1, whereas Macro F1 is the unweighted average over the nine-class label set and Weighted F1 weights each class by its evaluation support.

The baseline comparison exhibits a clear progression from generic inference to domain adaptation and task-specific training. Among the deterministic zero-shot baselines, BART-MNLI obtains an accuracy of 0.404, a Macro F1 of 0.353, and a Weighted F1 of 0.368. Its strong SUICIDAL\_THOUGHTS F1 (0.871) indicates that explicit high-risk language is readily captured, whereas its near-zero recall on WORTHLESSNESS (F1=0.029) reveals the limitations of independent generic entailment scores for a mutually exclusive, dataset-specific label space. The recent medical-domain MedGemma-4B-IT zero-shot baseline substantially improves over BART-MNLI, reaching 0.716 accuracy, 0.684 Macro F1, and 0.726 Weighted F1. The zero-shot MentalBERT cosine-similarity classifier, despite using no task-specific training, is a competitive reference point (0.704 accuracy, 0.731 Macro F1, 0.689 Weighted F1), performing particularly well on FATIGUE (F1=0.895) and SLEEP\_ISSUES (F1=0.921) but poorly on DEPRESSED\_MOOD (F1=0.504), where its definition-similarity signal is diluted by the category's broad and heterogeneous surface forms.

The trained encoder baselines demonstrate the importance of task-specific representation learning. DeBERTa reaches $0.760\pm0.035$ accuracy but a much lower Macro F1 of $0.547\pm0.025$, reflecting collapsed performance on the low-support APPETITE\_CHANGE and PSYCHOMOTOR categories (F1=0.000 in the representative seed). MentalRoBERTa is the strongest conventional encoder baseline, reaching $0.919\pm0.011$ accuracy, $0.878\pm0.010$ Macro F1, and $0.916\pm0.011$ Weighted F1. The proposed MentalSBERT-S reaches $0.934\pm0.002$ accuracy, $0.898\pm0.002$ Macro F1, and $0.932\pm0.002$ Weighted F1, exceeding MentalRoBERTa by 0.015, 0.020, and 0.016, respectively, with a visibly tighter seed-to-seed spread. To assess whether this gap reflects a stable effect rather than sampling variation, we ran a hierarchical bootstrap that jointly resamples the three training seeds and the test instances (10,000 resamples). The 95\% confidence interval for the Macro-F1 difference between MentalSBERT-S and MentalRoBERTa is $[-0.019,\,0.065]$ ($p=0.334$), so the numerical advantage of MentalSBERT-S over the strongest conventional encoder baseline is not statistically conclusive at this test-set scale. The specific contribution of the contrastive objective is examined separately below.

The generative baselines show that prompting alone remains behind supervised adaptation on this closed-set sentence-level task. DeepSeek zero-shot obtains 0.809 accuracy, 0.762 Macro F1, and 0.807 Weighted F1; few-shot prompting reaches 0.781, 0.758, and 0.793; and retrieval-augmented prompting reaches 0.762, 0.763, and 0.774. Directly fine-tuning DeepSeek-R1-Distill-Qwen-14B on the Stage-1 task improves on all three prompting variants, reaching $0.918\pm0.010$ accuracy, $0.891\pm0.020$ Macro F1, and $0.916\pm0.010$ Weighted F1, slightly below MentalSBERT-S on all three metrics. The fine-tuned generative model requires autoregressive generation for every sentence, whereas MentalSBERT-S requires none; given its comparable or better accuracy, this makes the lighter contrastively fine-tuned sentence encoder the more practical choice for the Stage-1 component, and it is used as such throughout the remainder of the paper.

MentalSBERT-S benefits from the joint classification and contrastive objective, though the gap is modest at this data scale. The BCE-only variant obtains an accuracy of $0.923\pm0.003$ and a Macro F1 of $0.881\pm0.024$, whereas the full BCE+triplet objective reaches $0.934\pm0.002$ and $0.898\pm0.002$, respectively -- a gain of 0.011 accuracy and 0.017 Macro F1, with a visibly tighter seed-to-seed spread on Macro F1. The largest per-class gain from adding the contrastive term is on PSYCHOMOTOR (F1 0.477 to 0.586), the lowest-support category, consistent with contrastive training providing the most benefit where classification supervision alone has the fewest positive examples to learn from. The training pairs use same-symptom anchor--positive pairs and random cross-symptom negatives. Within-category negatives are not used because the corpus does not identify semantically incompatible sentence pairs within a symptom label; treating unannotated same-label examples as negatives would oppose the BCE class supervision.

\begin{figure}[H]
\centering
\includegraphics[
width=\linewidth,
height=0.28\textheight,
keepaspectratio
]{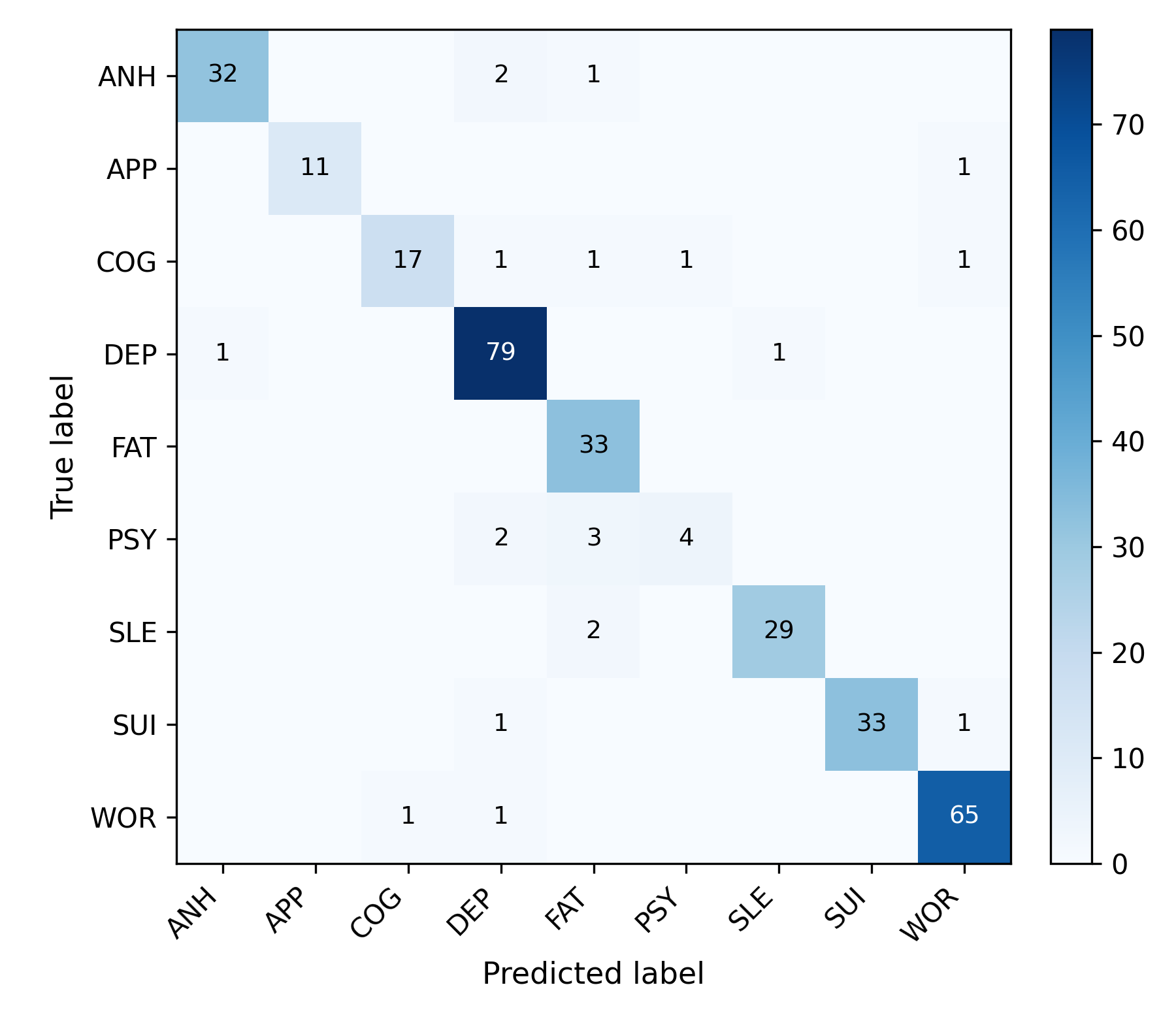}

\caption{Confusion matrix for MentalSBERT-S on the nine-class Stage-1 symptom-candidate task, representative seed 42. Rows denote the ground-truth DSM-5 symptom category and columns denote the category predicted by MentalSBERT-S on the 324-instance test partition; cell values are instance counts, with darker shading indicating a higher count. A perfectly accurate classifier would place all mass on the diagonal.}
\label{fig:confusion_matrix}
\end{figure}

The confusion matrix in Fig.~\ref{fig:confusion_matrix} presents the representative seed-42 result. Predictions are strongly concentrated along the diagonal, especially for DEPRESSED\_MOOD, FATIGUE, SUICIDAL\_THOUGHTS, and WORTHLESSNESS, indicating reliable recognition of several well-represented or semantically distinctive symptom categories. The remaining errors are concentrated in the lower-support PSYCHOMOTOR category, which is most often confused with the semantically related FATIGUE and DEPRESSED\_MOOD categories. This pattern is consistent with the per-class results in Tables~\ref{tab:stage1_prf_part1} and~\ref{tab:stage1_prf_part2} and indicates that the principal remaining difficulty lies in fine-grained boundary discrimination for rare categories rather than recognition of the major symptom classes.

The efficiency results in Table~\ref{tab:efficiency_comparison} highlight the practical cost of generative inference relative to the encoder approach. MentalSBERT-S classifies all 324 test sentences in under half a second of total inference time, several orders of magnitude faster than any generative baseline, while requiring no autoregressive generation. The DeepSeek and MedGemma-4B-IT prompting baselines and the fine-tuned DeepSeek model each require one generation per sentence and consume between $8.3\times10^{4}$ and $4.3\times10^{5}$ tokens in total, whereas MentalSBERT-S consumes under 6,000 tokens for the entire test partition. Combined with its comparable or better accuracy relative to the fine-tuned generative model, this efficiency advantage motivates using MentalSBERT-S as the Stage-1 component of the complete pipeline.

\subsection{Stage-2 Results Analysis}
\label{subsec:stage2_results}

Stage~2 evaluates whether a candidate symptom is \textsc{Present} or \textsc{Absent} in the target sentence. The proposed configuration fine-tunes DeepSeek-R1-Distill-Qwen-14B together with a concise candidate-specific, DSM-5-informed symptom definition. Within one structured inference call, the model first forms a binary judgment and then checks that judgment for consistency with the supplied definition. This procedure is evaluated as sentence-level annotation verification rather than as a complete clinical diagnostic assessment.

Tables~\ref{tab:stage2_prf_part1} and~\ref{tab:stage2_prf_part2} break down the proposed verifier's \textsc{Present}-class precision, recall, and F1 by candidate symptom category, complementing the aggregate results in Table~\ref{tab:stage2_overall_results}.

\begin{table*}[!t]
\centering
\captionsetup{justification=centering}
\caption{\textsc{Present}-class precision, recall, and F1-score by candidate symptom category on the Stage-2 verification task (multi-method comparison; cross-encoder and fine-tuned rows shown for representative seed 42) for ANH=ANHEDONIA; APP=APPETITE\_CHANGE; COG=COGNITIVE\_ISSUES; DEP=DEPRESSED\_MOOD; FAT=FATIGUE.}
\label{tab:stage2_prf_part1}
\footnotesize
\renewcommand{\arraystretch}{1.15}
\setlength{\tabcolsep}{3.8pt}
\resizebox{\textwidth}{!}{
\begin{tabular}{cccccc}
\toprule
\multicolumn{1}{c}{\textbf{Method}} &
\multicolumn{1}{c}{\textbf{ANH}} &
\multicolumn{1}{c}{\textbf{APP}} &
\multicolumn{1}{c}{\textbf{COG}} &
\multicolumn{1}{c}{\textbf{DEP}} &
\multicolumn{1}{c}{\textbf{FAT}} \\
\midrule
BART-MNLI zero-shot & \makecell{0.765/1.000/0.867} & \makecell{0.667/1.000/0.800} & \makecell{0.550/1.000/0.710} & \makecell{0.786/0.917/0.846} & \makecell{0.906/0.967/0.935} \\
MedGemma-4B-IT zero-shot & \makecell{0.743/1.000/0.852} & \makecell{0.667/1.000/0.800} & \makecell{0.550/1.000/0.710} & \makecell{0.794/0.900/0.844} & \makecell{0.909/1.000/0.952} \\
DeepSeek zero-shot & \makecell{0.870/0.769/0.816} & \makecell{0.700/0.875/0.778} & \makecell{0.526/0.909/0.667} & \makecell{0.872/0.683/0.766} & \makecell{0.906/0.967/0.935} \\
DeepSeek few-shot & \makecell{0.815/0.846/0.830} & \makecell{0.875/0.875/0.875} & \makecell{0.600/0.818/0.692} & \makecell{0.778/0.817/0.797} & \makecell{0.935/0.967/0.951} \\
DeepSeek RAG & \makecell{0.833/0.769/0.800} & \makecell{0.700/0.875/0.778} & \makecell{0.588/0.909/0.714} & \makecell{0.875/0.817/0.845} & \makecell{0.935/0.967/0.951} \\
DeBERTa-v3 cross-encoder & \makecell{0.743/1.000/0.852} & \makecell{0.667/1.000/0.800} & \makecell{0.524/1.000/0.688} & \makecell{0.741/1.000/0.851} & \makecell{0.909/1.000/0.952} \\
MentalSBERT cross-encoder & \makecell{0.769/0.769/0.769} & \makecell{0.667/1.000/0.800} & \makecell{0.556/0.909/0.690} & \makecell{0.738/0.983/0.843} & \makecell{0.909/1.000/0.952} \\
DeepSeek (fine-tuned, direct) & \makecell{0.833/0.769/0.800} & \makecell{0.700/0.875/0.778} & \makecell{0.588/0.909/0.714} & \makecell{0.824/0.933/0.875} & \makecell{0.906/0.967/0.935} \\
DeepSeek (fine-tuned, definition-only) & \makecell{0.852/0.885/0.868} & \makecell{0.700/0.875/0.778} & \makecell{0.526/0.909/0.667} & \makecell{0.818/0.900/0.857} & \makecell{0.909/1.000/0.952} \\
\textbf{DeepSeek (fine-tuned, proposed)} & \makecell{\textbf{0.808/0.808/0.808}} & \makecell{\textbf{0.667/0.750/0.706}} & \makecell{\textbf{0.556/0.909/0.690}} & \makecell{\textbf{0.831/0.900/0.864}} & \makecell{\textbf{0.909/1.000/0.952}} \\
\bottomrule
\end{tabular}}
\end{table*}

\begin{table*}[!t]
\centering
\captionsetup{justification=centering}
\caption{\textsc{Present}-class precision, recall, and F1-score by candidate symptom category on the Stage-2 verification task (multi-method comparison; cross-encoder and fine-tuned rows shown for representative seed 42) for PSY=PSYCHOMOTOR; SLE=SLEEP\_ISSUES; SUI=SUICIDAL\_THOUGHTS; WOR=WORTHLESSNESS.}
\label{tab:stage2_prf_part2}
\footnotesize
\renewcommand{\arraystretch}{1.15}
\setlength{\tabcolsep}{3.8pt}
\resizebox{0.75\textwidth}{!}{
\begin{tabular}{ccccc}
\toprule
\multicolumn{1}{c}{\textbf{Method}} &
\multicolumn{1}{c}{\textbf{PSY}} &
\multicolumn{1}{c}{\textbf{SLE}} &
\multicolumn{1}{c}{\textbf{SUI}} &
\multicolumn{1}{c}{\textbf{WOR}} \\
\midrule
BART-MNLI zero-shot & \makecell{0.667/1.000/0.800} & \makecell{0.581/1.000/0.735} & \makecell{0.970/0.941/0.955} & \makecell{0.815/0.863/0.838} \\
MedGemma-4B-IT zero-shot & \makecell{0.625/0.833/0.714} & \makecell{0.581/1.000/0.735} & \makecell{0.968/0.882/0.923} & \makecell{0.790/0.961/0.867} \\
DeepSeek zero-shot & \makecell{0.750/0.500/0.600} & \makecell{0.607/0.944/0.739} & \makecell{0.964/0.794/0.871} & \makecell{0.971/0.667/0.791} \\
DeepSeek few-shot & \makecell{1.000/0.333/0.500} & \makecell{0.542/0.722/0.619} & \makecell{1.000/0.853/0.921} & \makecell{0.875/0.824/0.848} \\
DeepSeek RAG & \makecell{0.667/0.333/0.444} & \makecell{0.552/0.889/0.681} & \makecell{0.968/0.882/0.923} & \makecell{0.857/0.824/0.840} \\
DeBERTa-v3 cross-encoder & \makecell{0.667/1.000/0.800} & \makecell{0.581/1.000/0.735} & \makecell{0.971/1.000/0.986} & \makecell{0.761/1.000/0.864} \\
MentalSBERT cross-encoder & \makecell{0.625/0.833/0.714} & \makecell{0.586/0.944/0.723} & \makecell{0.971/1.000/0.986} & \makecell{0.769/0.980/0.862} \\
DeepSeek (fine-tuned, direct) & \makecell{0.667/0.667/0.667} & \makecell{0.615/0.889/0.727} & \makecell{0.967/0.853/0.906} & \makecell{0.820/0.980/0.893} \\
DeepSeek (fine-tuned, definition-only) & \makecell{0.667/1.000/0.800} & \makecell{0.593/0.889/0.711} & \makecell{0.969/0.912/0.939} & \makecell{0.814/0.941/0.873} \\
\textbf{DeepSeek (fine-tuned, proposed)} & \makecell{\textbf{0.571/0.667/0.615}} & \makecell{\textbf{0.654/0.944/0.773}} & \makecell{\textbf{0.971/0.971/0.971}} & \makecell{\textbf{0.860/0.961/0.907}} \\
\bottomrule
\end{tabular}}
\end{table*}

\begin{table*}[!t]
\centering
\caption{Overall Stage-2 performance on the 324-instance test partition. Values for trained methods report mean $\pm$ sample SD over seeds 42, 52, and 62. \textit{Note:} Macro F1 is the unweighted mean over the \textsc{Present} and \textsc{Absent} classes; Weighted F1 is support-weighted. Deterministic zero-shot, few-shot, and retrieval-augmented baselines are reported once.}
\label{tab:stage2_overall_results}
\scriptsize
\setlength{\tabcolsep}{4pt}
\resizebox{\textwidth}{!}{
\begin{tabular}{lccccc}
\toprule
\textbf{Method} & \textbf{Accuracy} & \textbf{Macro F1} & \textbf{Weighted F1} & \textbf{Absent F1} & \textbf{Present F1} \\
\midrule
BART-MNLI zero-shot & 0.750 & 0.553 & 0.703 & 0.257 & 0.850 \\
MedGemma-4B-IT zero-shot & 0.747 & 0.530 & 0.692 & 0.212 & 0.849 \\
DeepSeek zero-shot & 0.710 & 0.636 & 0.719 & 0.472 & 0.800 \\
DeepSeek few-shot & 0.731 & 0.634 & 0.730 & 0.446 & 0.823 \\
DeepSeek RAG & 0.741 & 0.642 & 0.737 & 0.455 & 0.830 \\
DeBERTa-v3 cross-encoder & $0.755\pm0.004$ & $0.488\pm0.102$ & $0.675\pm0.048$ & $0.120\pm0.208$ & $0.857\pm0.004$ \\
MentalSBERT cross-encoder & $0.750\pm0.011$ & $0.524\pm0.040$ & $0.690\pm0.018$ & $0.197\pm0.084$ & $0.852\pm0.009$ \\
DeepSeek (fine-tuned, direct) & $0.766\pm0.011$ & $0.656\pm0.019$ & $0.754\pm0.010$ & $0.461\pm0.040$ & $0.851\pm0.010$ \\
DeepSeek (fine-tuned, definition-only) & $0.770\pm0.010$ & $0.630\pm0.052$ & $0.745\pm0.026$ & $0.404\pm0.102$ & $0.857\pm0.001$ \\
\textbf{DeepSeek (fine-tuned, proposed)} & $\mathbf{0.780\pm0.005}$ & $\mathbf{0.658\pm0.021}$ & $\mathbf{0.761\pm0.007}$ & $0.453\pm0.048$ & $\mathbf{0.862\pm0.007}$ \\
\bottomrule
\end{tabular}}
\end{table*}

\begin{table*}[!t]
\centering
\caption{Efficiency comparison across the retained Stage-2 baselines and ablations. Values report mean$\pm$sample SD where seed-specific measurements are available.}
\label{tab:stage2_efficiency}
\scriptsize
\resizebox{\textwidth}{!}{
\begin{tabular}{ccccc}
\toprule
\multicolumn{1}{c}{\textbf{Method}} &
\multicolumn{1}{c}{\textbf{Input tokens/sample}} &
\multicolumn{1}{c}{\textbf{Output tokens/sample}} &
\multicolumn{1}{c}{\textbf{Latency/sample (s)}} &
\multicolumn{1}{c}{\textbf{Throughput (samples/s)}} \\
\midrule
BART-MNLI zero-shot & 225.3 & -- & 0.010 & 101.71 \\
MedGemma-4B-IT zero-shot & 466.5 & 8.1 & 0.288 & 3.48 \\
DeepSeek zero-shot & 440.6 & 345.0 & 12.059 & 0.08 \\
DeepSeek few-shot & 1382.1 & 207.0 & 7.320 & 0.14 \\
DeepSeek RAG & 686.7 & 316.7 & 11.056 & 0.09 \\
DeBERTa-v3 cross-encoder & $232.3\pm0.0$ & -- & $0.0054\pm0.0000$ & 184.43 \\
MentalSBERT cross-encoder & $235.4\pm0.0$ & -- & $0.0039\pm0.0000$ & 252.45 \\
DeepSeek (fine-tuned, direct) & $194.8\pm0.0$ & $6.1\pm0.2$ & $0.491\pm0.016$ & 2.04 \\
DeepSeek (fine-tuned, definition-only) & $234.5\pm0.0$ & $35.7\pm0.8$ & $2.641\pm0.033$ & 0.38 \\
\textbf{DeepSeek (fine-tuned, proposed)} & $\mathbf{253.5\pm0.0}$ & $\mathbf{43.5\pm1.0}$ & $\mathbf{3.066\pm0.015}$ & \textbf{0.33} \\
\bottomrule
\end{tabular}}
\end{table*}

Table~\ref{tab:stage2_overall_results} compares the proposed verifier with prompting-based LLMs, zero-shot NLI and medical-LLM baselines, and supervised discriminative cross-encoders. Among the deterministic and prompted baselines, the DeepSeek zero-shot, few-shot, and RAG configurations reach the highest Macro F1 (0.636, 0.634, and 0.642, respectively) but the lowest accuracy (0.710--0.741) of any baseline family, since their higher Absent F1 (0.446--0.472) comes at the cost of frequently rejecting genuinely \textsc{Present} candidates. BART-MNLI and MedGemma-4B-IT zero-shot achieve higher accuracy (0.750 and 0.747) but substantially lower Absent F1 (0.257 and 0.212), revealing a strong tendency to accept the supplied candidate. The trained cross-encoders show a similar imbalance: both DeBERTa-v3 and MentalSBERT cross-encoders reach approximately 0.75 accuracy and Present F1 above 0.85, but DeBERTa-v3 is markedly unstable across seeds (Macro F1 $0.488\pm0.102$, with two of three seeds collapsing to predicting \textsc{Present} for nearly every instance), while the MentalSBERT cross-encoder is more consistent but still weak on the minority \textsc{Absent} class ($0.197\pm0.084$).
\begin{figure}[H]
\centering
\includegraphics[width=\columnwidth,keepaspectratio]{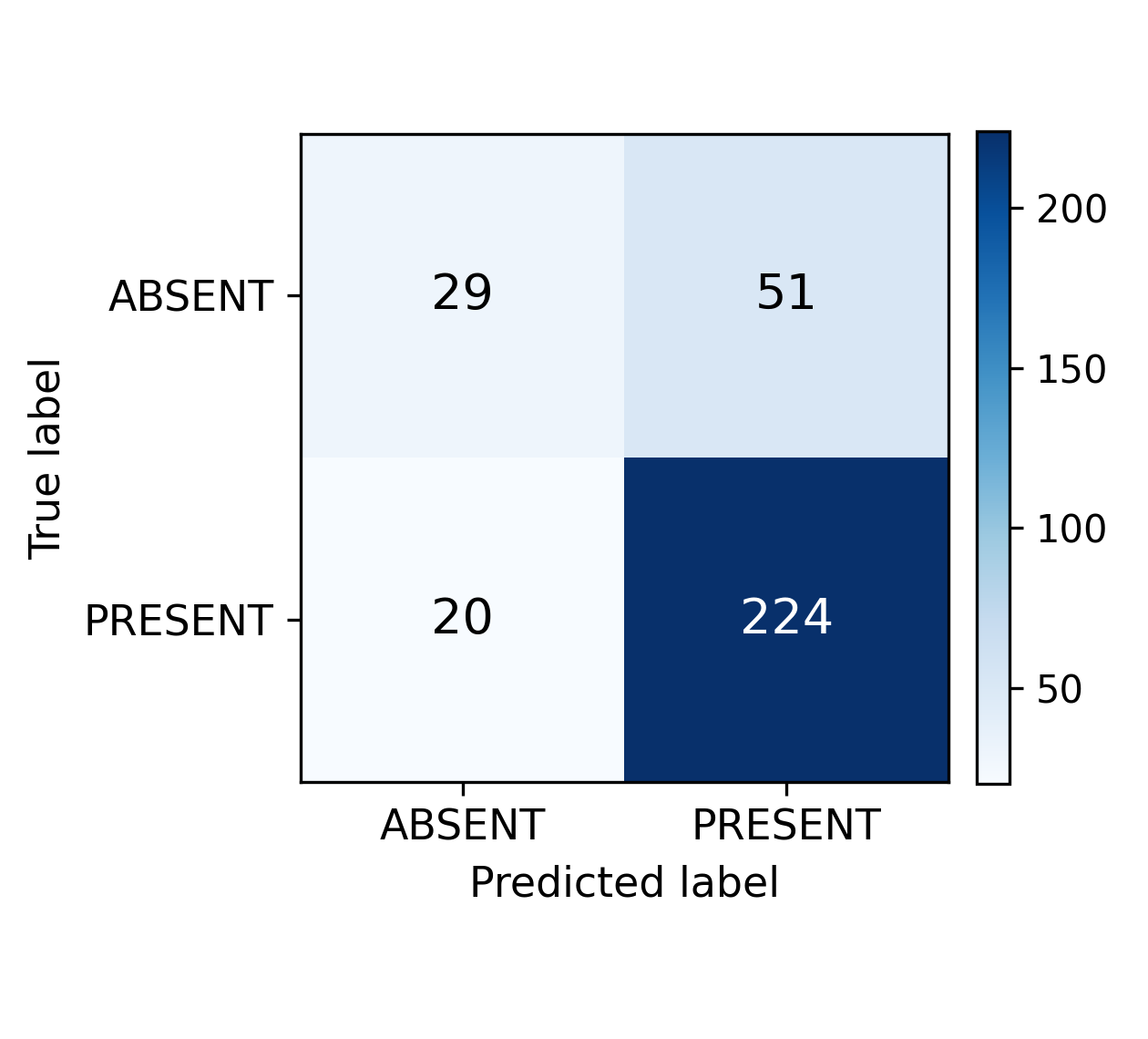}
\caption{Confusion matrix for the proposed Stage-2 verifier on the binary \textsc{Present}/\textsc{Absent} candidate-verification task, representative seed 42. Rows denote the ground-truth status of the Stage-1 candidate and columns denote the status predicted by the verifier on the 324-instance test partition (244 \textsc{Present}, 80 \textsc{Absent}); cell values are instance counts.}
\label{fig:stage2_confusion_matrix}
\end{figure}

Fine-tuning DeepSeek directly on the verification task without definitions or self-checking (\textit{direct}) already outperforms all zero-shot and cross-encoder baselines, reaching $0.766\pm0.011$ accuracy and $0.656\pm0.019$ Macro F1. Adding only the DSM-5 definition (\textit{definition-only}) lowers Macro F1 to $0.630\pm0.052$, mainly because Absent F1 drops from $0.461\pm0.040$ to $0.404\pm0.102$, suggesting that the definition alone may bias the model toward accepting candidates rather than rejecting unsupported ones. Adding the self-check step on top of the definition restores Macro F1 to $0.658\pm0.021$ and yields the best accuracy, $0.780\pm0.005$. A hierarchical bootstrap jointly resampling the three training seeds and test instances (10,000 resamples) found no significant pairwise differences: direct vs.\ definition-only, 95\% CI $[-0.073,\,0.118]$ ($p=0.687$); definition-only vs.\ the proposed self-check configuration, $[-0.122,\,0.072]$ ($p=0.627$); and direct vs.\ proposed, $[-0.071,\,0.064]$ ($p=0.970$). Thus, the observed pattern---definition-only reducing Macro F1 and self-checking recovering and slightly exceeding the direct setting---should be interpreted as a suggestive trend rather than a statistically established effect.

To compare the proposed verifier with the strongest external baseline, we evaluated it against DeepSeek RAG, the best-performing prompting configuration. The bootstrap 95\% CI for the Macro-F1 difference is $[-0.055,\,0.082]$ ($p=0.665$), indicating that the fine-tuned verifier's numerical advantage is not statistically conclusive at the current test-set scale, despite higher accuracy (0.780 vs.\ 0.741) and Present F1 (0.862 vs.\ 0.830).

Figure~\ref{fig:stage2_confusion_matrix} shows that the verifier correctly identifies 224 of 244 \textsc{Present} and 29 of 80 \textsc{Absent} instances. The 51 false accepts versus 20 false rejects is consistent with the lower Absent F1 in Table~\ref{tab:stage2_overall_results} and the class imbalance. Tables~\ref{tab:stage2_prf_part1} and~\ref{tab:stage2_prf_part2} show further category-level variation: F1 is highest for FATIGUE and SUICIDAL\_THOUGHTS (0.952 and 0.971) and lowest for PSYCHOMOTOR and COGNITIVE\_ISSUES (0.615 and 0.690), indicating greater reliability for well-supported, semantically clearer categories and similar rare-category difficulty to Stage~1. Table~\ref{tab:stage2_efficiency} shows that cross-encoders remain three to four orders of magnitude faster than generative verifiers. Among fine-tuned DeepSeek variants, the direct model is fastest ($0.491\pm0.016$s/sample) because it outputs only a binary decision, while definition-only and self-check generate short rationales ($35.7\pm0.8$ and $43.5\pm1.0$ tokens), increasing latency to $2.641\pm0.033$s and $3.066\pm0.015$s per sample.

Overall, the Stage-2 results support the definition-plus-self-check design: it achieves the highest accuracy and Present F1, while all fine-tuned models---direct, definition-only, and self-check---outperform all zero-shot, few-shot, and cross-encoder baselines on accuracy. Differences among the fine-tuned variants, however, are not statistically conclusive at the current test-set scale and should be interpreted as trends rather than settled conclusions. As in Stage~1, the remaining errors, especially for the sparse \textsc{Absent} class, indicate that the verifier is a sentence-level annotation component rather than a substitute for clinical diagnosis.
\subsection{Two-Stage Inference Result Analysis}

\begin{table*}[!t]
\centering
\captionsetup{justification=centering}
\caption{Per-class precision, recall, and F1-score on the final ten-class end-to-end evaluation (multi-seed methods and the proposed pipeline shown for representative seed 42) for ANH=ANHEDONIA; APP=APPETITE\_CHANGE; COG=COGNITIVE\_ISSUES; DEP=DEPRESSED\_MOOD; FAT=FATIGUE.}
\label{tab:e2e_prf_part1}
\footnotesize
\renewcommand{\arraystretch}{1.15}
\setlength{\tabcolsep}{3.8pt}
\resizebox{\textwidth}{!}{
\begin{tabular}{cccccc}
\toprule
\multicolumn{1}{c}{\textbf{Method}} &
\multicolumn{1}{c}{\textbf{ANH}} &
\multicolumn{1}{c}{\textbf{APP}} &
\multicolumn{1}{c}{\textbf{COG}} &
\multicolumn{1}{c}{\textbf{DEP}} &
\multicolumn{1}{c}{\textbf{FAT}} \\
\midrule
BART-MNLI zero-shot & \makecell{0.361/0.846/0.506} & \makecell{0.500/0.375/0.429} & \makecell{0.242/0.727/0.364} & \makecell{0.471/0.667/0.552} & \makecell{0.366/0.867/0.515} \\
MentalBERT zero-shot & \makecell{0.543/0.731/0.623} & \makecell{0.667/1.000/0.800} & \makecell{0.333/0.818/0.474} & \makecell{0.724/0.350/0.472} & \makecell{0.824/0.933/0.875} \\
MedGemma-4B-IT zero-shot & \makecell{0.679/0.731/0.704} & \makecell{0.636/0.875/0.737} & \makecell{0.333/0.636/0.438} & \makecell{0.444/0.733/0.553} & \makecell{0.750/0.500/0.600} \\
DeBERTa & \makecell{0.000/0.000/0.000} & \makecell{0.000/0.000/0.000} & \makecell{0.000/0.000/0.000} & \makecell{0.649/0.800/0.716} & \makecell{0.789/0.500/0.612} \\
DeepSeek zero-shot & \makecell{0.500/0.538/0.519} & \makecell{0.778/0.875/0.824} & \makecell{0.412/0.636/0.500} & \makecell{0.528/0.633/0.576} & \makecell{0.774/0.800/0.787} \\
MentalRoBERTa & \makecell{0.643/0.692/0.667} & \makecell{0.727/1.000/0.842} & \makecell{0.833/0.455/0.588} & \makecell{0.708/0.850/0.773} & \makecell{0.789/1.000/0.882} \\
DeepSeek few-shot & \makecell{0.606/0.769/0.678} & \makecell{0.750/0.750/0.750} & \makecell{0.471/0.727/0.571} & \makecell{0.724/0.700/0.712} & \makecell{0.744/0.967/0.841} \\
DeepSeek RAG & \makecell{0.600/0.808/0.689} & \makecell{0.714/0.625/0.667} & \makecell{0.500/0.909/0.645} & \makecell{0.623/0.717/0.667} & \makecell{0.871/0.900/0.885} \\
Single-stage supervised DeepSeek QLoRA & \makecell{0.720/0.692/0.706} & \makecell{0.778/0.875/0.824} & \makecell{0.400/0.545/0.462} & \makecell{0.718/0.850/0.779} & \makecell{0.789/1.000/0.882} \\
\textbf{Proposed two-stage pipeline} & \makecell{\textbf{0.750/0.692/0.720}} & \makecell{\textbf{0.625/0.625/0.625}} & \makecell{\textbf{0.438/0.636/0.519}} & \makecell{\textbf{0.743/0.867/0.800}} & \makecell{\textbf{0.769/1.000/0.870}} \\
\bottomrule
\end{tabular}}
\end{table*}

\begin{table*}[!t]
\centering
\captionsetup{justification=centering}
\caption{Per-class precision, recall, and F1-score on the final ten-class end-to-end evaluation (multi-seed methods and the proposed pipeline shown for representative seed 42) for PSY=PSYCHOMOTOR; SLE=SLEEP\_ISSUES; SUI=SUICIDAL\_THOUGHTS; WOR=WORTHLESSNESS; NO=\texttt{NO\_SYMPTOM}.}
\label{tab:e2e_prf_part2}
\footnotesize
\renewcommand{\arraystretch}{1.15}
\setlength{\tabcolsep}{3.8pt}
\resizebox{\textwidth}{!}{
\begin{tabular}{cccccc}
\toprule
\multicolumn{1}{c}{\textbf{Method}} &
\multicolumn{1}{c}{\textbf{PSY}} &
\multicolumn{1}{c}{\textbf{SLE}} &
\multicolumn{1}{c}{\textbf{SUI}} &
\multicolumn{1}{c}{\textbf{WOR}} &
\multicolumn{1}{c}{\textbf{NO}} \\
\midrule
BART-MNLI zero-shot & \makecell{0.067/0.167/0.095} & \makecell{0.333/0.056/0.095} & \makecell{0.935/0.853/0.892} & \makecell{0.917/0.216/0.349} & \makecell{0.429/0.037/0.069} \\
MentalBERT zero-shot & \makecell{0.250/0.167/0.200} & \makecell{0.531/0.944/0.680} & \makecell{0.589/0.971/0.733} & \makecell{0.434/0.647/0.520} & \makecell{0.474/0.113/0.182} \\
MedGemma-4B-IT zero-shot & \makecell{0.100/0.333/0.154} & \makecell{0.556/0.833/0.667} & \makecell{0.897/0.765/0.825} & \makecell{0.816/0.608/0.697} & \makecell{0.548/0.212/0.306} \\
DeBERTa & \makecell{0.000/0.000/0.000} & \makecell{0.000/0.000/0.000} & \makecell{0.681/0.941/0.790} & \makecell{0.750/0.882/0.811} & \makecell{0.387/0.600/0.471} \\
DeepSeek zero-shot & \makecell{0.143/0.167/0.154} & \makecell{0.548/0.944/0.694} & \makecell{0.912/0.912/0.912} & \makecell{0.754/0.843/0.796} & \makecell{0.526/0.250/0.339} \\
MentalRoBERTa & \makecell{0.000/0.000/0.000} & \makecell{0.593/0.889/0.711} & \makecell{0.941/0.941/0.941} & \makecell{0.793/0.902/0.844} & \makecell{0.480/0.300/0.369} \\
DeepSeek few-shot & \makecell{0.600/0.500/0.545} & \makecell{0.552/0.889/0.681} & \makecell{0.967/0.853/0.906} & \makecell{0.677/0.824/0.743} & \makecell{0.605/0.325/0.423} \\
DeepSeek RAG & \makecell{0.375/0.500/0.429} & \makecell{0.567/0.944/0.708} & \makecell{0.968/0.882/0.923} & \makecell{0.854/0.804/0.828} & \makecell{0.556/0.312/0.400} \\
Single-stage supervised DeepSeek QLoRA & \makecell{0.200/0.167/0.182} & \makecell{0.667/0.667/0.667} & \makecell{0.935/0.853/0.892} & \makecell{0.815/0.863/0.838} & \makecell{0.500/0.362/0.420} \\
\textbf{Proposed two-stage pipeline} & \makecell{\textbf{0.400/0.333/0.364}} & \makecell{\textbf{0.680/0.944/0.791}} & \makecell{\textbf{0.969/0.912/0.939}} & \makecell{\textbf{0.842/0.941/0.889}} & \makecell{\textbf{0.604/0.362/0.453}} \\
\bottomrule
\end{tabular}}
\end{table*}

\begin{table*}[!t]
\centering
\captionsetup{justification=centering}
\caption{Overall performance in the final ten-class end-to-end evaluation on the 324-instance test partition. Trained methods report mean $\pm$ sample SD over seeds 42, 52, and 62; deterministic single-run baselines are reported once.}
\label{tab:two_stage_overall_results}
\scriptsize
\renewcommand{\arraystretch}{1.12}
\setlength{\tabcolsep}{4pt}
\resizebox{\textwidth}{!}{
\begin{tabular}{lccc}
\toprule
\multicolumn{1}{c}{\textbf{Method}} &
\multicolumn{1}{c}{\textbf{Accuracy}} &
\multicolumn{1}{c}{\textbf{Macro F1}} &
\multicolumn{1}{c}{\textbf{Weighted F1}} \\
\midrule
BART-MNLI zero-shot & 0.444 & 0.387 & 0.386 \\
MentalBERT zero-shot & 0.549 & 0.556 & 0.499 \\
MedGemma-4B-IT zero-shot & 0.565 & 0.568 & 0.559 \\
DeBERTa & $0.484\pm0.093$ & $0.258\pm0.081$ & $0.412\pm0.102$ \\
DeepSeek zero-shot & 0.623 & 0.610 & 0.605 \\
MentalRoBERTa & $0.691\pm0.016$ & $0.633\pm0.026$ & $0.663\pm0.016$ \\
DeepSeek few-shot & 0.682 & 0.685 & 0.666 \\
DeepSeek RAG & 0.685 & 0.684 & 0.672 \\
Single-stage supervised DeepSeek QLoRA & $0.698\pm0.003$ & $0.661\pm0.015$ & $0.684\pm0.004$ \\
\textbf{Proposed two-stage pipeline} & $\mathbf{0.731\pm0.005}$ & $\mathbf{0.690\pm0.011}$ & $\mathbf{0.715\pm0.010}$ \\
\bottomrule
\end{tabular}}
\end{table*}

\begin{table*}[!t]\centering\captionsetup{justification=centering}
\caption{Efficiency comparison for the final ten-class end-to-end evaluation on the 324-instance test partition.\newline\footnotesize\textit{Note:} LLM Gen is the number of autoregressive generations. The proposed pipeline's cost is the sum of the Stage-1 encoder pass and the Stage-2 verifier call for every test sentence. Multi-seed configurations report mean$\pm$SD across seeds.}
\label{tab:e2e_efficiency}\scriptsize\renewcommand{\arraystretch}{1.08}\setlength{\tabcolsep}{3pt}
\resizebox{\textwidth}{!}{\begin{tabular}{cccccccc}\toprule\multicolumn{1}{c}{\textbf{Method}} &
\multicolumn{1}{c}{\textbf{LLM Gen}} &
\multicolumn{1}{c}{\textbf{Total Time (s)}} &
\multicolumn{1}{c}{\textbf{Avg Latency (s)}} &
\multicolumn{1}{c}{\textbf{Throughput}} &
\multicolumn{1}{c}{\textbf{Avg Input Tok.}} &
\multicolumn{1}{c}{\textbf{Avg Output Tok.}} &
\multicolumn{1}{c}{\textbf{Total Tokens}} \\\midrule
BART-MNLI zero-shot & 0 & 5.76 & 0.018 & 56.25 & 286.6 & 0.0 & 92874 \\
MentalBERT zero-shot & 0 & 0.20 & 0.001 & 1651.61 & 18.3 & 0.0 & 5923 \\
DeBERTa & 0 & $1.05\pm0.00$ & $0.003\pm0.000$ & $308.69\pm1.32$ & $276.1\pm0.0$ & $0.0\pm0.0$ & $89463\pm0$ \\
MentalRoBERTa & 0 & $0.41\pm0.01$ & $0.001\pm0.000$ & $790.29\pm25.93$ & $290.3\pm0.0$ & $0.0\pm0.0$ & $94067\pm0$ \\
MedGemma-4B-IT zero-shot & 324 & 155.60 & 0.480 & 2.08 & 407.7 & 14.6 & 136824 \\
DeepSeek zero-shot & 324 & 1184.28 & 3.655 & 0.27 & 544.7 & 93.1 & 206652 \\
DeepSeek few-shot & 324 & 1417.77 & 4.376 & 0.23 & 1517.9 & 109.7 & 527348 \\
DeepSeek RAG & 324 & 1286.30 & 3.970 & 0.25 & 830.9 & 100.1 & 301671 \\
Single-stage supervised DeepSeek QLoRA & 324 & $459.03\pm0.22$ & $1.417\pm0.001$ & $0.71\pm0.00$ & $495.7\pm0.0$ & $18.7\pm0.1$ & $166674\pm40$ \\
\textbf{Proposed two-stage pipeline} & \textbf{324} & $\mathbf{978.32\pm30.25}$ & $\mathbf{3.020\pm0.093}$ & $\mathbf{0.33\pm0.01}$ & $\mathbf{271.5\pm0.0}$ & $\mathbf{43.4\pm0.9}$ & $\mathbf{102041\pm294}$ \\
\bottomrule\end{tabular}}
\end{table*}
The final experiment evaluates the complete framework without access to gold symptom labels. For each target sentence, Stage~1 (MentalSBERT-S) proposes a symptom candidate, and Stage~2 (the fine-tuned DeepSeek verifier) then determines whether that candidate is \textsc{Present} or \textsc{Absent}. A \textsc{Present} decision preserves the Stage-1 symptom, whereas an \textsc{Absent} decision produces \texttt{NO\_SYMPTOM}. The resulting label space contains the nine core symptom categories plus \texttt{NO\_SYMPTOM}. This genuinely end-to-end evaluation differs from the isolated Stage-1 and Stage-2 results reported above in that Stage-2 verifies whatever category Stage-1 actually predicted, including its errors, rather than the gold category.

Table~\ref{tab:two_stage_overall_results} reports aggregate performance, comparing the proposed pipeline against encoder, NLI-style, medical-LLM, and direct-prompting baselines, together with a directly supervised single-stage DeepSeek-R1-Distill-Qwen-14B QLoRA baseline trained end-to-end on the same ten-class task and training partition. All trained methods are reported as mean$\pm$sample SD over seeds 42, 52, and 62; deterministic single-run baselines are reported once.

Table~\ref{tab:e2e_efficiency} shows that the proposed pipeline's inference cost is dominated by the Stage-2 generation call: the Stage-1 MentalSBERT-S pass adds under half a second in total across all 324 test sentences, so the pipeline's average latency ($3.020\pm0.093$s/sample) closely tracks the isolated Stage-2 verifier cost reported in Table~\ref{tab:stage2_efficiency}. The proposed pipeline is markedly cheaper than every DeepSeek prompting baseline (zero-shot, few-shot, and RAG all exceed 3.6s/sample and consume $2$--$5\times$ more total tokens). It is slower per sample than the single-stage supervised DeepSeek QLoRA baseline ($1.417\pm0.001$s), since the single-stage classifier emits only a short class label in a single call whereas the proposed pipeline performs two model calls per sentence and generates a supporting rationale as part of the Stage-2 verification; despite this, the proposed pipeline consumes fewer total tokens overall ($102041\pm294$ versus $166674\pm40$) because its Stage-1 pass requires no generation at all. This latency-versus-accuracy trade-off is consistent with the accuracy advantage of the two-stage decomposition reported in Table~\ref{tab:two_stage_overall_results}.

Among the baselines, accuracy and Macro F1 improve steadily from generic zero-shot inference (BART-MNLI: 0.444/0.387) through domain zero-shot and encoder methods (MentalBERT: 0.549/0.556; MedGemma zero-shot: 0.565/0.568; DeBERTa: $0.484\pm0.093$/$0.258\pm0.081$, unstable across seeds) to task-adapted encoders and prompted generation (MentalRoBERTa: $0.691\pm0.016$/$0.633\pm0.026$; DeepSeek few-shot: 0.682/0.685; DeepSeek RAG: 0.685/0.684). Directly fine-tuning DeepSeek-R1-Distill-Qwen-14B as a single-stage ten-class classifier improves further, reaching $0.698\pm0.003$ accuracy and $0.661\pm0.015$ Macro F1, and outperforms every zero-shot, few-shot, RAG, and encoder baseline.

The proposed two-stage pipeline -- MentalSBERT-S for candidate generation followed by the fine-tuned DeepSeek verifier -- reaches $0.731\pm0.005$ accuracy, $0.690\pm0.011$ Macro F1, and $0.715\pm0.010$ Weighted F1, the best result of any evaluated method. Because the single-stage supervised baseline uses the identical backbone, training partition, and comparable QLoRA optimisation settings, the comparison against it isolates the benefit of decomposing the task into candidate generation and candidate-specific verification rather than any difference in model capacity or training data. A hierarchical bootstrap that jointly resamples the three training seeds and the test instances (10,000 resamples) gives a 95\% CI of $[-0.053,\,0.105]$ for the Macro-F1 difference between the two-stage pipeline and the single-stage baseline ($p=0.498$), so the numerical advantage of the two-stage decomposition, while consistent in direction across all three seeds, is not statistically conclusive at this test-set scale.
\begin{figure}[H]
\centering
\includegraphics[width=\columnwidth]{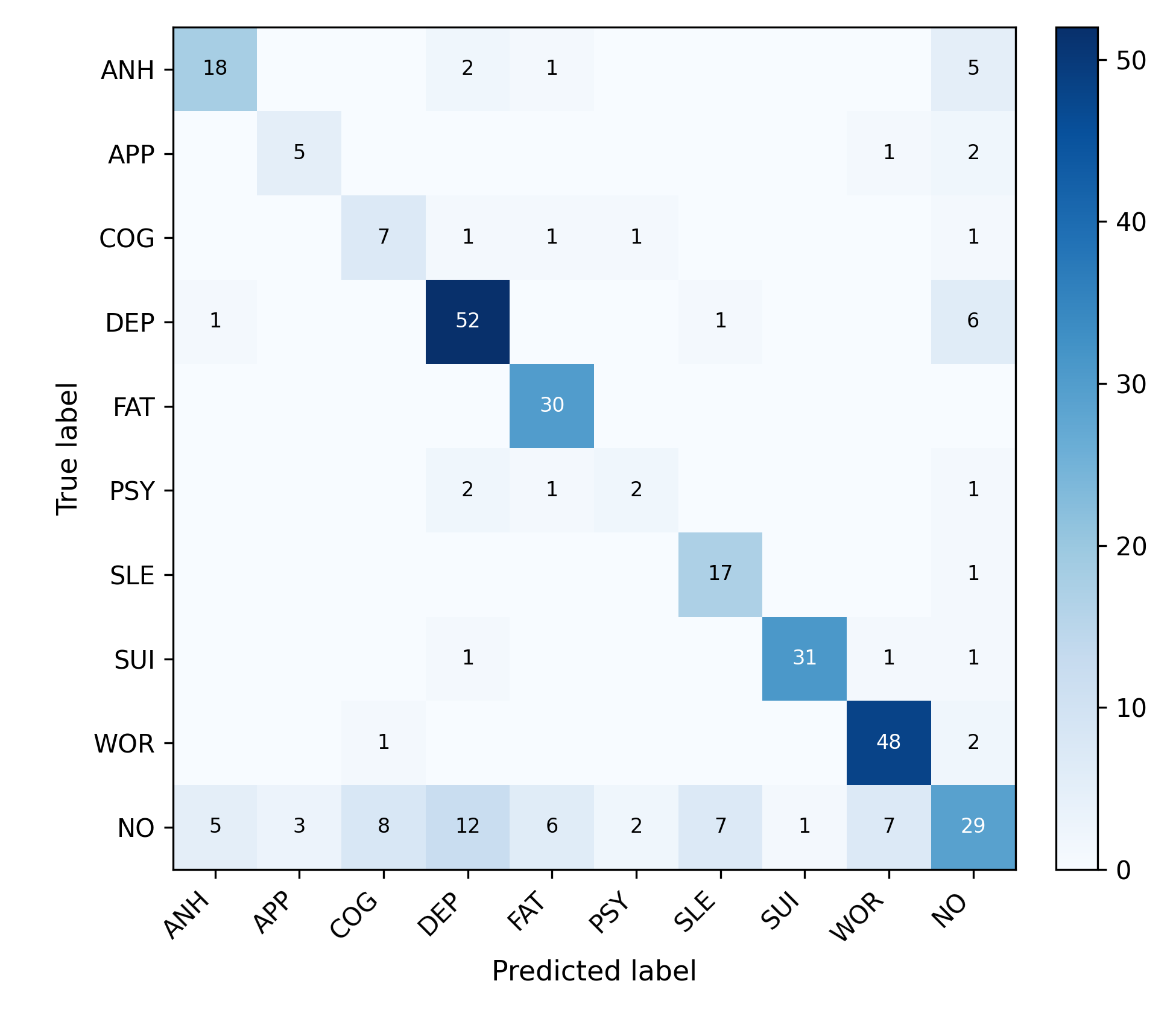}
\caption{Confusion matrix for the complete two-stage pipeline on the final ten-class end-to-end task (nine DSM-5 symptom categories plus \texttt{NO\_SYMPTOM}), representative seed 42 (accuracy 0.738, Macro F1 0.697). Rows denote the ground-truth ten-class label and columns denote the label produced by the full pipeline (Stage-1 candidate generation followed by Stage-2 verification) on the 324-instance test partition; cell values are instance counts.}
\label{fig:fig5_multiclass_confusion_matrix}
\end{figure}
Tables~\ref{tab:e2e_prf_part1} and~\ref{tab:e2e_prf_part2}, together with Fig.~\ref{fig:fig5_multiclass_confusion_matrix}, show the representative seed-42 run (accuracy 0.738, Macro F1 0.697, Weighted F1 0.721) in per-class detail. Predictions are strongly concentrated along the diagonal for SUICIDAL\_THOUGHTS (F1 0.939), WORTHLESSNESS (0.889), FATIGUE (0.870), and DEPRESSED\_MOOD (0.800). The weakest categories are the low-support PSYCHOMOTOR (F1 0.364, 6 test instances) and \texttt{NO\_SYMPTOM} (F1 0.453), the latter reflecting the accumulation of both Stage-1 candidate errors and Stage-2 verification errors: a sentence is only correctly mapped to \texttt{NO\_SYMPTOM} when Stage-2 correctly rejects whatever candidate Stage-1 proposed, so \texttt{NO\_SYMPTOM} recall is bounded by Stage-2's Absent-class sensitivity reported in Table~\ref{tab:stage2_overall_results}. This end-to-end error compounding -- an incorrect Stage-1 candidate that Stage-2 accepts becomes an unrecoverable error, since Stage-2 can only accept or reject the proposed candidate rather than substitute a different one -- is an inherent limitation of the candidate-generation-then-verification decomposition.

We additionally assessed semantic agreement between generated Stage-2 rationales and the expert-authored annotation rationales provided by ReDSM-5 using Sentence-BERT embeddings~\cite{b30}, restricted to the subset of test instances where both the Stage-1 candidate and the Stage-2 decision were jointly correct. Across seeds 42, 52, and 62, mean cosine similarity was $0.597\pm0.003$, with $3.5\%\pm0.3\%$ of rationales exceeding the prespecified high-similarity threshold of 0.75. This embedding-based measure quantifies lexical-semantic agreement with the expert-authored rationales and should not be interpreted as an independent clinical-correctness score.

A consultant psychiatrist additionally conducted a single-rater blinded audit of 32 model-generated explanations produced by the final sanitized Stage-2 verifier (Table~\ref{tab:clinician_explanation_audit}). The audit packet was resampled to address a limitation of an earlier pilot round: rather than conditioning on correctly classified outputs only, the 32 audited cases are stratified into four equal-sized subsets crossing the verifier's predicted status (\textsc{Present}/\textsc{Absent}) with whether that prediction was correct, so that rationale quality could be assessed separately for correct and incorrect verifier decisions rather than only for cases the system already got right. As in the pilot round, the rater was blinded to the gold label and to correctness, and the results are reported here as a preliminary, single-rater assessment by a coauthor rather than an independent external clinical evaluation. Because the four subsets were sampled in equal number (eight each) rather than in proportion to their true frequency in the test partition (64.8\%/20.4\%/9.6\%/5.2\% for \textsc{Present}-correct/\textsc{Present}-incorrect/\textsc{Absent}-correct/\textsc{Absent}-incorrect, respectively), the overall score reported in Table~\ref{tab:clinician_explanation_audit} is computed as a population-weighted average of the four subset means using these true proportions as weights, rather than an unweighted average over the 32 rated cases; an unweighted average would overrepresent the rare \textsc{Absent} subsets and underrepresent the dominant \textsc{Present}-correct subset relative to the system's actual output distribution.

The population-weighted overall scores are $3.15$ for DSM-5 alignment and $2.52$ for clinical usefulness. The dominant factor is whether the verifier's decision was correct, not whether it was \textsc{Present} or \textsc{Absent}: correct decisions (74.4\% population weight) score $3.52$/$2.73$ on average, whereas incorrect decisions (25.6\% population weight) drop to $2.11$/$1.90$. This confirms the concern that motivated resampling the audit -- rationale quality for incorrect verifier decisions is substantially lower than for correct ones, so an audit restricted to correct cases alone overestimates practical explanation quality. In contrast, the gap between \textsc{Present} and \textsc{Absent} rationales is much smaller once both correct and incorrect decisions are included ($3.11$/$2.54$ for \textsc{Present} versus $3.40$/$2.40$ for \textsc{Absent}, population-weighted), unlike the larger \textsc{Present}-versus-\textsc{Absent} gap observed in the smaller, correctness-conditioned pilot round. The rater's written comments further indicate that some \textsc{Present}-incorrect cases reflect an upstream Stage-1 candidate error (e.g., a sentence describing a suicide attempt routed to Stage-2 under an unrelated candidate symptom rather than \texttt{SUICIDAL\_THOUGHTS}) rather than a Stage-2 rationale failure per se, and that in at least one case the rater's clinical judgment diverged from the underlying ReDSM-5 gold annotation itself, which we note as a limitation of the source annotations rather than of the model.

\begin{table}[!t]
\centering
\caption{Single-rater blinded psychiatrist audit of 32 model-generated explanations, stratified by predicted status and correctness. Both DSM-5 alignment and clinical usefulness are rated on a five-point Likert scale (1=lowest; 5=highest). Subset rows report the sample mean$\pm$SD over the eight equally-sampled cases in that subset. The Overall row is a population-weighted average of the four subset means, weighted by each subset's true frequency in the 324-instance test partition (64.8\%/20.4\%/9.6\%/5.2\%) rather than by the equal 25\% sampling weight used to construct the audit.}
\label{tab:clinician_explanation_audit}
\scriptsize
\renewcommand{\arraystretch}{1.10}
\setlength{\tabcolsep}{4pt}
\begin{tabular}{ccc}
\toprule
\multicolumn{1}{c}{\textbf{Subset}} &
\multicolumn{1}{c}{\textbf{DSM-5 alignment}} &
\multicolumn{1}{c}{\textbf{Clinical usefulness}} \\
\midrule
\textsc{Present}, correct ($n=8$, weight 64.8\%) & $3.50\pm1.20$ & $2.75\pm0.89$ \\
\textsc{Present}, incorrect ($n=8$, weight 20.4\%) & $1.88\pm0.99$ & $1.88\pm0.83$ \\
\textsc{Absent}, correct ($n=8$, weight 9.6\%) & $3.62\pm1.30$ & $2.62\pm0.92$ \\
\textsc{Absent}, incorrect ($n=8$, weight 5.2\%) & $3.00\pm1.69$ & $2.00\pm0.93$ \\
\midrule
\textbf{Overall (population-weighted, $n=32$)} & $\mathbf{3.15}$ & $\mathbf{2.52}$ \\
\bottomrule
\end{tabular}
\end{table}

Overall, the end-to-end results support the two-stage decomposition: the proposed pipeline reaches the best accuracy, Macro F1, and Weighted F1 of any evaluated method, including a matched single-stage supervised baseline, though the margin over that baseline is not yet statistically conclusive at the current test-set scale.

\section{Discussion}
\label{sec:discussion}

This study develops a two-stage framework for sentence-level DSM-5 symptom recognition that separates symptom candidate generation from candidate-specific verification. Stage~1 uses a contrastively fine-tuned sentence encoder, MentalSBERT-S, to generate an initial symptom candidate for every sentence. Stage~2 then evaluates whether that candidate is \textsc{Present} or \textsc{Absent} using the target sentence, its post context, and a candidate-specific DSM-5-informed definition, combined with a self-check step that verifies the preliminary judgment against the supplied definition before answering. This decomposition differs from direct end-to-end prediction because the two stages address distinct questions: which symptom is the most plausible candidate, and whether the available evidence is sufficient to retain that candidate. The resulting outputs should be interpreted as sentence-level research annotations rather than psychiatric diagnoses.

Two design choices account for most of the improvement observed in this study. First, task-specific fine-tuning provides a decisive and consistent gain over prompting-based inference at both stages: the fine-tuned Stage-1 encoder and Stage-2 verifier outperform every zero-shot, few-shot, and retrieval-augmented baseline by a wide margin, confirming that adapting model parameters to this annotation scheme, rather than relying on prompting alone, is central to the framework's accuracy. Second, decomposing the task into candidate generation and definition-grounded verification is architecturally well motivated and is the best-performing configuration among all evaluated methods, including every single-stage alternative, even though its numerical margin over a matched single-stage supervised baseline does not reach statistical significance at the current test-set scale (discussed below).

The Stage-1 results reported above show that MentalSBERT-S matches the accuracy of the directly fine-tuned generative alternative (DeepSeek) while requiring no autoregressive generation, making it a substantially cheaper choice for the candidate-generation role without a measurable accuracy cost. The Stage-2 ablation isolates the contribution of the two design choices layered onto direct verification: adding the DSM-5 definition alone numerically reduces Macro F1 relative to the direct baseline, while adding the self-check step on top of the definition recovers and slightly exceeds the direct-only level; none of the pairwise differences reaches significance at the current test-set scale, so this is reported as a trend rather than a settled effect.

The end-to-end comparison against a matched single-stage supervised DeepSeek QLoRA baseline is the most direct test of whether the two-stage decomposition itself is beneficial, since both methods share the same backbone, training partition, and comparable optimisation settings. The two-stage pipeline outperforms the single-stage baseline on accuracy, Macro F1, and Weighted F1, consistently across all three seeds, though the hierarchical bootstrap comparison does not reach significance at $p<0.05$. We report this honestly as a consistent numerical trend rather than a proven effect, and note that a larger evaluation partition would be needed to determine whether the gap is a genuine, stable property of the decomposition. This accuracy advantage comes at a latency cost: the two-stage pipeline is slower per sample than the single-stage baseline (Table~\ref{tab:e2e_efficiency}), since it performs two model calls and generates a supporting rationale rather than a single short class label, though it remains markedly cheaper than any prompting-based baseline and consumes fewer total tokens overall.

The error analysis reveals an architectural limitation shared by both single-stage and two-stage designs at the current data scale: rare categories such as PSYCHOMOTOR remain difficult across every evaluated method (Tables~\ref{tab:stage2_prf_part1}--\ref{tab:stage2_prf_part2} and~\ref{tab:e2e_prf_part1}--\ref{tab:e2e_prf_part2}), and the two-stage pipeline's \texttt{NO\_SYMPTOM} performance is bounded by Stage-2's sensitivity to the \textsc{Absent} class. Because Stage-2 can only accept or reject the single candidate Stage-1 proposes, an incorrect Stage-1 candidate that Stage-2 accepts becomes an unrecoverable end-to-end error; Stage-2 has no mechanism to substitute a different symptom category. A possible extension would allow Stage-2 to compare a small set of Stage-1 candidates or request a revision, but such a mechanism constitutes a distinct architecture and would require separate validation.

The explanation-quality evidence should be read with its limitations in mind. The Sentence-BERT similarity measure quantifies lexical-semantic overlap with expert-authored rationales, not clinical correctness. The psychiatrist audit, while a useful preliminary signal, is rated by a single clinician who is also a coauthor, and is therefore better described as a preliminary single-rater audit than an independent external clinical evaluation. Its stratified, population-weighted design shows that rationale quality is governed primarily by whether the verifier's decision is correct (population-weighted DSM-5 alignment/clinical usefulness of $3.52$/$2.73$ for correct decisions versus $2.11$/$1.90$ for incorrect ones) rather than by whether the candidate is \textsc{Present} or \textsc{Absent} ($3.11$/$2.54$ versus $3.40$/$2.40$), which both confirms the concern that motivated resampling the audit and revises the earlier, correctness-conditioned finding that \textsc{Present} rationales were the primary weakness.

Several additional limitations apply to the study as a whole. The dataset is small and long-tailed, and several rare categories have wide seed-to-seed variation despite multi-seed evaluation; the observed class distribution reflects the sampled social-media corpus and should not be interpreted as clinically representative prevalence. The source annotations encode sentence-level symptom relevance and do not provide sufficient information to establish DSM-5 duration, functional impairment, or full episode-level diagnostic criteria, so the system's outputs are sentence-level evidence markers rather than diagnostic judgments. Finally, the framework does not reconcile multiple sentence-level decisions into a post-level diagnosis; cross-sentence consistency, temporal evidence, symptom co-occurrence, and post-level aggregation remain outside its validated scope.

A related limitation concerns statistical power. None of the pairwise comparisons reported in this study -- the Stage-2 ablation among direct, definition-only, and self-check variants, the proposed verifier against the strongest prompting baseline, and the end-to-end two-stage pipeline against the matched single-stage baseline -- reached significance at the conventional $p<0.05$ threshold under the hierarchical bootstrap. This reflects the joint resampling of both training seeds and the 324-instance test partition (and, for several per-symptom breakdowns, even smaller category-level subsets), a considerably stricter standard than a seed-only comparison that ignores instance-level sampling uncertainty. The observed effect sizes are consistently small (one to three percentage points in Macro F1) relative to the sampling noise inherent to a few-hundred-instance test set, so a substantially larger evaluation partition would be needed to establish or rule out these effects with confidence. We regard the contribution of this work as resting on convergent, non-p-value evidence rather than any single significance test: the proposed configuration achieves the best point estimate on every aggregate metric among all evaluated methods, consistently across all three seeds and in the direction predicted by its architectural motivation -- definition-grounded self-checking reduces spurious symptom acceptance, and decomposing candidate generation from verification separates two distinct error sources -- while requiring substantially less inference cost than every prompting-based alternative. 

Overall, the results support decomposing sentence-level symptom recognition into candidate generation and definition-grounded candidate verification. The two-stage pipeline improves over strong encoder, NLI-style, medical-LLM, direct-prompting, and matched single-stage supervised alternatives under a shared evaluation protocol, while the seed-level and error analyses make its remaining limitations explicit. The evidence supports the framework as a research method for producing sentence-anchored symptom hypotheses and rationales; it does not establish a diagnostic system or a substitute for professional clinical assessment.

\section{Conclusion}

This study proposes a two-stage framework for sentence-level DSM-5 depression symptom recognition, combining a contrastively fine-tuned encoder for symptom-candidate generation with a definition-grounded, self-checking verifier for candidate-specific \textsc{Present}/\textsc{Absent} judgment. The framework separates two tasks: identifying the most plausible symptom candidate and determining whether the available evidence is sufficient to retain it.

Stage~1 uses MentalSBERT-S to generate symptom-aware representations and select an initial candidate without autoregressive generation. Stage~2 uses a fine-tuned DeepSeek verifier with the target sentence, post context, and a candidate-specific DSM-5-informed definition, checking its preliminary decision against that definition. \textsc{Present} retains the Stage-1 candidate, while \textsc{Absent} maps the sentence to \texttt{NO\_SYMPTOM}.

End-to-end, the framework achieved $0.731\pm0.005$ accuracy, $0.690\pm0.011$ Macro F1, and $0.715\pm0.010$ Weighted F1 across three seeds, outperforming all evaluated baselines, including a matched single-stage supervised DeepSeek QLoRA model, although this margin was not statistically significant. The two-stage design incurs additional latency but remains substantially cheaper than prompting-based baselines. Generated rationales reached $0.597\pm0.003$ Sentence-BERT similarity with expert rationales on jointly correct predictions. A preliminary psychiatrist audit gave population-weighted DSM-5 alignment and clinical usefulness scores of $3.15$ and $2.52$, with rationale quality strongly dependent on decision correctness.

Limitations include the small, long-tailed dataset, ambiguous symptom expressions, dependence on Stage~1 candidates, limited statistical power, and the preliminary clinical audit. The system is intended for sentence-level research annotation and explanation, not diagnosis or replacement of professional clinical assessment.

\section*{Conflicting Interests}
The authors declare that there is no conflict of interest.

\section*{Funding}
This work was supported by LARSyS funding
DOI: 10.54499/LA/P/0083/2020
and DOI: 10.54499/UID/50009/
2025.

\section*{Ethical Approval}
This research follows ethical principles and guidelines applicable to IEEE Access publications. The study performs secondary analysis of ReDSM-5, a publicly released, de-identified dataset of Reddit posts and sentence-level annotations \cite{b18}. Although the underlying text originates from human authors, this work involves no direct interaction with human participants, no collection of new personal data, and no identifiable information: all posts and annotations were obtained from ReDSM-5 in its already-anonymized, publicly distributed form and used strictly under its stated licence and usage terms. Under the Ethics Committee Regulation of Instituto Superior Técnico, University of Lisbon, ethics-committee review applies to research involving human beings directly as research subjects or indirectly susceptible to being affected by it, informed consent, or the protection of privacy and personal data; this study falls outside that scope, as it neither collects new data nor involves any direct or indirect interaction with participants. On this basis, and consistent with Committee on Publication Ethics (COPE) guidance on secondary analysis of pre-existing anonymized public datasets, the authors determined that a new ethics committee review at Instituto Superior Técnico, University of Lisbon, was not required for this study. The study adheres to publication ethics guidelines based on principles upheld by the Committee on COPE.

\section*{Guarantor}
Weiming Li.

\section*{Contributorship}
Weiming Li conducted the literature review, conceptualized
the study, performed data collection and analysis, and wrote
the manuscript. Catarina Barata and João Sanches reviewed
the manuscript and provided valuable revision suggestions.
Miguel Constante contributed psychiatric expertise and re
viewed the DSM-5-related clinical consistency of the pro
posed framework. All authors reviewed and approved the
final version of the manuscript.

\section*{Acknowledgment}
The authors would like to acknowledge Instituto Superior Técnico, University of Lisbon, the Institute for Systems and Robotics (ISR), and LARSyS for providing the research environment and infrastructure supporting this work.

\appendices

\section{Training Configurations and Prompt Templates}
\label{app:training_prompts}

This appendix reports the principal training settings, task definitions, prompt templates, and decoding constraints used in the final experiments. All reported evaluation results were obtained on the fixed evaluation partition. The evaluation labels and reference explanations were used only for performance and explanation-quality assessment and were not included in model prompts or used for checkpoint optimisation.

\subsection{Software Environment}

\begin{table}[H]
\centering
\caption{Principal software environment used for model training and evaluation, listing the exact package versions of the retained \texttt{fine\_tuning\_deepseek} environment, identical across all three random-seed runs, to support exact reproduction of the reported results.}
\label{tab:software_environment}
\footnotesize
\renewcommand{\arraystretch}{1.08}
\begin{tabular}{cc}
\toprule
\multicolumn{1}{c}{\textbf{Package}} &
\multicolumn{1}{c}{\textbf{Version}} \\
\midrule
Python & 3.10.18 \\
PyTorch & 2.10.0.dev20250922+cu128 \\
Transformers & 4.56.2 \\
PEFT & 0.17.1 \\
bitsandbytes & 0.48.1 \\
NumPy & 1.26.4 \\
scikit-learn & 1.7.2 \\
\bottomrule
\end{tabular}
\end{table}

The package versions were read from the retained \texttt{fine\_tuning\_deepseek} environment used for the final experiments. All three random-seed runs used the same software environment and model identifiers.

\subsection{MentalSBERT-S Training Configuration}

MentalSBERT-S serves as the Stage-1 sentence encoder and symptom-candidate classifier. It is based on \texttt{sentence-transformers/all-mpnet-base-v2} and contains nine independent one-vs-rest sigmoid heads, one per depressive symptom category. Table~\ref{tab:mentalsbert_config} reports its principal training settings.

\begin{table}[H]
\centering
\caption{Training configuration of MentalSBERT-S, the Stage-1 nine-head sentence encoder used for symptom-candidate generation, including the base encoder, joint BCE-plus-triplet training objective, and optimisation hyperparameters shared across the three random-seed runs.}
\label{tab:mentalsbert_config}
\renewcommand{\arraystretch}{1.08}
\setlength{\tabcolsep}{5pt}
\footnotesize
\begin{tabular}{cc}
\toprule
\multicolumn{1}{c}{\textbf{Component}} &
\multicolumn{1}{c}{\textbf{Configuration}} \\
\midrule
Base encoder & \texttt{all-mpnet-base-v2} \\
Architecture & Nine-head one-vs-rest classifier \\
Output activation & Independent sigmoid heads \\
Training objective & BCE + triplet metric-learning loss \\
Optimizer & AdamW \\
Learning rate & $2\times10^{-5}$ \\
Batch size & 16 \\
Epochs & 5 \\
Warmup ratio & 0.10 \\
Weight decay & 0.01 \\
Triplet margin & 0.30 \\
Triplet-loss weight & 0.50 \\
Gradient clipping & 1.0 \\
Scheduler & Linear warmup and linear decay \\
Training seeds & 42, 52, and 62 \\
\bottomrule
\end{tabular}
\end{table}

\subsection{DeepSeek Verifier Fine-Tuning}

The Stage-2 verifier is initialised from DeepSeek-R1-Distill-Qwen-14B and adapted using 4-bit QLoRA for candidate-specific \textsc{Present}/\textsc{Absent} verification. Table~\ref{tab:deepseek_config} reports the fine-tuning configuration, and Table~\ref{tab:headB_task} summarises the task-specific input/output specification.

\begin{table}[H]
\centering
\caption{Fine-tuning configuration of the Stage-2 DeepSeek verifier, including the 4-bit QLoRA setup, target modules, and optimisation hyperparameters applied to \texttt{DeepSeek-R1-Distill-Qwen-14B} across the three random-seed runs.}
\label{tab:deepseek_config}
\renewcommand{\arraystretch}{1.08}
\setlength{\tabcolsep}{5pt}
\footnotesize
\begin{tabular}{cc}
\toprule
\multicolumn{1}{c}{\textbf{Item}} &
\multicolumn{1}{c}{\textbf{Configuration}} \\
\midrule
Base model & \texttt{DeepSeek-R1-Distill-Qwen-14B} \\
Model scale & 14B parameters \\
Maximum training length & 2048 tokens \\
Fine-tuning method & 4-bit QLoRA \\
Quantisation & NF4 with double quantisation \\
Compute precision & bfloat16 \\
LoRA rank $r$ & 16 \\
LoRA scaling $\alpha$ & 32 \\
LoRA dropout & 0.05 \\
Target modules & \texttt{q\_proj}, \texttt{k\_proj}, \texttt{v\_proj}, \texttt{o\_proj}, \\
 & \texttt{gate\_proj}, \texttt{up\_proj}, \texttt{down\_proj} \\
Epochs & 3 \\
Per-device batch size & 1 \\
Gradient accumulation & 8 \\
Effective batch size & 8 \\
Optimizer & AdamW \\
Learning rate & $2\times10^{-4}$ \\
Warmup ratio & 0.03 \\
Weight decay & 0.01 \\
Training seeds & 42, 52, and 62 \\
\bottomrule
\end{tabular}
\end{table}

\begin{table}[H]
\centering
\caption{Task-specific input/output specification of the Stage-2 verifier, describing the information supplied to the model at inference time and the structured fields it must produce, including the model-generated rationale used in the explanation-quality audit of Section~\ref{sec:discussion}.}
\label{tab:headB_task}
\renewcommand{\arraystretch}{1.08}
\setlength{\tabcolsep}{4pt}
\footnotesize
\begin{tabular}{>{\centering\arraybackslash}p{2.6cm}>{\centering\arraybackslash}p{5.4cm}}
\toprule
\multicolumn{1}{c}{\textbf{Component}} &
\multicolumn{1}{c}{\textbf{Configuration}} \\
\midrule
Task & Candidate-specific \textsc{Present}/\textsc{Absent} verification \\
Primary input & Stage-1 candidate symptom, target sentence, post context, and candidate-specific DSM-5-informed definition \\
Training objective & Causal-LM supervised fine-tuning \\
Prediction output & \texttt{self\_label}, \texttt{final\_label}, and \texttt{verify\_action} \\
Additional output & Model-generated rationale \\
Inference calls & One main call per sample; repair only after invalid JSON \\
\bottomrule
\end{tabular}
\end{table}

Each training instance pairs one target sentence with one candidate symptom and its \textsc{Present}/\textsc{Absent} status, following the candidate-specific annotation structure of ReDSM-5 (Section~\ref{subsec:dataset}); no additional negative candidates are synthesised. The 1,163-instance training partition and 280-instance development partition are used directly, without oversampling or class reweighting. The resulting label distribution is imbalanced toward \textsc{Present}: 904 \textsc{Present} versus 259 \textsc{Absent} instances in training (77.7\%/22.3\%) and 212 versus 68 in development (75.7\%/24.3\%), closely matching the 244/80 (75.3\%/24.7\%) split observed in the test partition. This imbalance motivates reporting Absent F1 and Present F1 separately throughout Section~\ref{subsec:stage2_results} rather than accuracy alone.

\subsection{Candidate-Specific DSM-5-Informed Definitions}

The following task-specific, DSM-5-informed definitions are supplied to the verifier according to the Stage-1 candidate:

\begin{itemize}
\item \textbf{ANHEDONIA:} reduced interest or inability to experience pleasure in normally enjoyable activities.
\item \textbf{APPETITE\_CHANGE:} a noticeable increase or decrease in appetite or weight not explained by dieting or medical conditions.
\item \textbf{COGNITIVE\_ISSUES:} difficulty thinking, concentrating, or making decisions, including mental fog or slowed thinking.
\item \textbf{DEPRESSED\_MOOD:} feelings of sadness, emptiness, hopelessness, or emotional heaviness.
\item \textbf{FATIGUE:} reduced energy, tiredness, or physical or mental exhaustion affecting functioning.
\item \textbf{PSYCHOMOTOR:} observable slowing or agitation in movement, restlessness, or difficulty initiating physical actions.
\item \textbf{SLEEP\_ISSUES:} difficulty falling asleep, remaining asleep, or excessive sleep, including disrupted or unrefreshing sleep.
\item \textbf{SUICIDAL\_THOUGHTS:} thoughts of death, suicidal ideas, wishing to be dead, or considering self-harm.
\item \textbf{WORTHLESSNESS:} strong negative self-evaluation, worthlessness, guilt, or harsh self-blame.
\end{itemize}

No deterministic keyword override, symptom-specific fallback, or gold-label fallback is applied at inference time.

\subsection{Stage-2 Verifier Prompt Template}

For each Stage-1 candidate, the verifier receives the target sentence, post context, and corresponding DSM-5-informed definition, then forms a binary judgment and checks it against the definition within the same generation call.

\begin{lstlisting}[basicstyle=\ttfamily\scriptsize]
You are a DSM-5 depression sentence annotation assistant.

Decide whether TARGET_SENTENCE expresses the requested label S
as PRESENT (1) or ABSENT (0).

Use TARGET_SENTENCE as the primary evidence, CONTEXT_POST as
supporting local context, and DSM5_GUIDELINE as the definition
of label S.

Important:
- This is sentence-level annotation, not a formal diagnosis.
- Do not require duration, severity, recurrence, functional
  impairment, or a complete diagnostic history.
- Brief, mild, implicit, or short expressions can be PRESENT
  when the target sentence or local context meaningfully
  expresses S.
- Do not use dataset-specific rules or superficial keyword
  matching.
- Do not mention annotations, gold labels, training data, or
  the technical decision process in the explanation.

[SYMPTOM]
{stage1_candidate}

[DSM5_GUIDELINE]
{candidate_specific_definition}

[TARGET_SENTENCE]
{sentence_text}

[CONTEXT_POST]
{context_post}

Step 1: INITIAL SENTENCE-LEVEL JUDGMENT
Set self_label=1 if the target sentence or local context
meaningfully expresses S; otherwise set self_label=0.

Step 2: DSM-5 CONSISTENCY CHECK
Conservatively verify self_label against DSM5_GUIDELINE.
Use KEEP_1 or KEEP_0 when the initial judgment is retained,
and CORRECT_TO_0 or CORRECT_TO_1 when it is corrected.

Step 3: EXPLANATION
Write a DSM-5-style explanation supporting final_label without
referring to an original annotation, gold label, training data,
or technical processing.

Output one JSON object and nothing else:
{
  "self_label": 0 or 1,
  "final_label": 0 or 1,
  "verify_action": "KEEP_0" or "CORRECT_TO_1" or
                   "KEEP_1" or "CORRECT_TO_0",
  "reason": "3--6 sentences supporting final_label"
}
\end{lstlisting}

\subsection{Inference Summary}

\begin{table}[H]
\centering
\caption{Decoding and generation settings used for the final Stage-2 verifier at inference time, applied identically across all reported test-partition evaluations and the clinician explanation audit.}
\label{tab:inference_config}
\renewcommand{\arraystretch}{1.08}
\setlength{\tabcolsep}{4pt}
\footnotesize
\begin{tabular}{>{\centering\arraybackslash}p{3.2cm}>{\centering\arraybackslash}p{3.9cm}}
\toprule
\multicolumn{1}{c}{\textbf{Setting}} &
\multicolumn{1}{c}{\textbf{Configuration}} \\
\midrule
Invocation & Every Stage-1 candidate \\
Temperature & Not applicable \\
Sampling & Disabled \\
Top-$p$ & 1.0 \\
Maximum new tokens & 192 \\
Primary prediction & Binary \texttt{final\_label} \\
Context use & Supporting local evidence \\
Additional generation & JSON repair only if required \\
Gold-label fallback & Disabled \\
\bottomrule
\end{tabular}
\end{table}

\subsection{Pipeline Output Example}

An example output generated by the proposed two-stage framework is shown below.

\begin{quote}
\textbf{Candidate symptom:} \\
\texttt{WORTHLESSNESS}

\vspace{0.4em}

\textbf{System status:} \\
\texttt{ABSENT}

\vspace{0.4em}

\textbf{Target sentence:} \\
\textit{I'm already happy with who I am and what I do.}

\vspace{0.4em}

\textbf{Supporting post context:} \\
\textit{What do you mean by upgrading myself? I'm already happy with who I am and what I do.}

\vspace{0.4em}

\textbf{Model-generated explanation:} \\
\textit{The sentence expresses a positive self-perception and satisfaction with life, which is not indicative of worthlessness.}
\end{quote}

\bibliographystyle{IEEEtran}
\bibliography{refs}

\vspace{-1.2cm}
\begin{IEEEbiography}[{\includegraphics[width=1in,height=1.25in,clip,keepaspectratio]{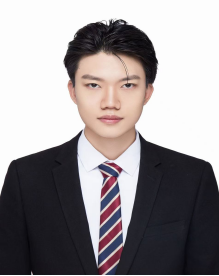}}]{WEIMING LI } received the bachelor’s degree from Tiangong University, China, and the master’s degree from the University of Leeds, U.K. He is currently pursuing the Ph.D. degree with the Institute for Systems and Robotics (ISR), LARSyS, Instituto Superior Técnico, University of Lisbon. His research interests include natural language processing and artificial intelligence.
\end{IEEEbiography}

\vspace{-1.2cm}

\begin{IEEEbiography}[{\includegraphics[width=1in,height=1.25in,clip,keepaspectratio]{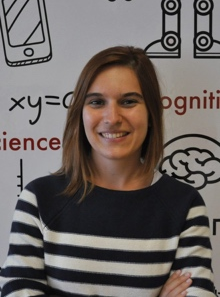}}]{Catarina Barata}  is an Assistant Professor in the Department of Electrical Engineering at Instituto Superior Técnico (IST) and Researcher at the Institute of Systems and Robotics of Lisbon (ISR-Lisbon). She is a member of the European Laboratory for Learning and Intelligent Systems (ELLIS) and its Lisbon Unit (Lisbon). Her main line of research focuses on the application of machine learning models to image
analysis problems, with a focus on the development of explainable artificial intelligence models. In 2021 she received a Google Research Award for her work in personalizing treatment for cancer patients. She participated in several national and European research projects. She has published over 20 articles in scientific journals and presented more than 40 papers at international conferences.
\end{IEEEbiography}

\vspace{-1.2cm}

\begin{IEEEbiography}[{\includegraphics[width=1in,height=1.25in,clip,keepaspectratio]{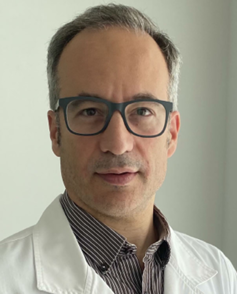}}]{MIGUEL CONSTANTE} is a Consultant General Adult Psychiatrist, he completed his specialist training in the U.K, at the Maudsley Hospital, acquiring membership of the Royal College of Psychiatrists and his Ph.D. degree in neurophysiology of psychosis from the Institute of Psychiatry, KCL. He is currently based in Lisbon, providing clinical care and managing a broad range of psychiatric disorders in varied clinical settings at ULSLOD (public) and Luz Saúde (private). Simultaneously, he is an invited Professor at Catolica Medical School. He is actively involved in medical teaching, psychiatry training, and service development. His research interests include exploring, alongside biomedical engineers, digital psychiatry interventions.
\end{IEEEbiography}

\vspace{-2.5cm}

\begin{IEEEbiography}[{\includegraphics[width=1in,height=1.25in,clip,keepaspectratio]{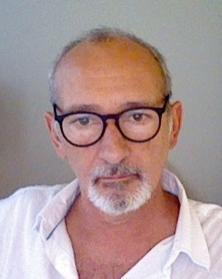}}]{João Sanches} is Full Professor at the Department of Bioengineering, Instituto Superior Técnico (IST), Universidade de Lisboa (UL) where he coordinates the Master and Doctoral programs in Biomedical Engineering. He’s senior research at Institute for Systems and Robotics / LARSyS where he conducts research in medical and biological image analysis and statistical signal processing of biomedical data, mainly, physiological and behavioral data. He is Senior Member of the IEEE Engineering in Medicine and Biology Society (EMBS). As a group leader, he has fostered a multidisciplinary environment that brings together experts from biology, medicine, engineering, and computer science. This collaboration has resulted in approximately 200 international publications.

\end{IEEEbiography}

\EOD

\end{document}